\documentclass[preprint,12pt]{elsarticle}

\usepackage{amsmath,amssymb,amsfonts}
\usepackage{algorithm}
\usepackage{algpseudocode}
\usepackage{graphicx}
\usepackage{textcomp}
\usepackage{xcolor}
\usepackage{hyperref}
\usepackage{booktabs}
\usepackage{multirow}
\usepackage{array}
\usepackage{caption}
\usepackage{subcaption}
\usepackage{float}
\usepackage{makecell}
\usepackage{enumitem}
\usepackage{tabularx}
\usepackage{adjustbox}
\usepackage{rotating}
\usepackage{threeparttable}
\usepackage{stfloats}
\usepackage{placeins}

\hypersetup{
    colorlinks=true,
    linkcolor=blue,
    filecolor=magenta,
    urlcolor=cyan,
}

\algrenewcommand{\algorithmicrequire}{\textbf{Input:}}
\algrenewcommand{\algorithmicensure}{\textbf{Output:}}

\makeatletter
\renewcommand{\ALG@beginalgorithmic}{\setcounter{ALG@line}{1}}
\makeatother

\begin{document}

\begin{frontmatter}

\title{Federated Temporal Attention Intelligence for Cyber-Resilient IoMT: Lightweight Digital Twins and PPO-Driven Honeypot Deception}

\author[label1]{Syed Zeeshan Haider}
\ead{zeeshan.haider.syed1010@gmail.com}

\author[label1]{Anwar Shah\corref{cor1}}
\ead{anwar.shah@nu.edu.pk}

\author[label1]{Muneeb Arif}
\ead{muneebrajpoot9797@gmail.com}

\author[label1]{Hamza Iftikhar}
\ead{hamzaiftikharbhatti@gmail.com}

\author[label1]{Waqas Ali}
\ead{waqas.ali@nu.edu.pk}

\cortext[cor1]{Corresponding author}

\address[label1]{FAST National University of Computer and Emerging Sciences, Islamabad, Pakistan}

\begin{abstract}
The rapid proliferation of Internet of Medical Things (IoMT) devices introduces critical cybersecurity vulnerabilities in healthcare environments where resource-constrained medical devices operate under strict latency requirements and stringent data-privacy regulations. This paper presents the Lightweight Digital Twin and Federated Reinforcement Learning (LDT-FRL) framework, a privacy-preserving defense architecture integrating four complementary mechanisms: a Temporal Attention Encoder (TAE) built on a GRU backbone with learned temporal self-attention for flow-level threat classification; lightweight LSTM-based Digital Twins trained on normal-class traffic to generate per-device anomaly scores that gate the TAE classifier through a learned sigmoid coupling; a Federated Proximal Policy Optimization (PPO) agent selecting among ALLOW, ISOLATE, and HONEYPOT\_REDIRECT actions based on a seven-dimensional state capturing classifier confidence, entropy, anomaly magnitude, and traffic composition; and an intelligent honeypot layer that converts redirected suspicious traffic into actionable threat intelligence. A federated aggregation strategy is employed, using exponential moving average (EMA)-smoothed per-client validation losses as inverse-weighted FedAvg coefficients. The purpose of this strategy is a stabilization of global model updates under non-IID client distributions. An evaluation on CICDDoS~2019 (431,371 raw records, 215,674 temporal windows, 61 features, 16 attack classes), shows that the framework achieves 99.66\% test accuracy  together with a macro-F1 score of 0.9913 by a federated round~9. This represents a convergence that is 81\% faster than the DTFL-CD baseline while matching its minimum CPU utilisation of 44\% against that same baseline. A second evaluation on TON-IoT (211,043 raw records, 70,335 windows, 16 features, 10 classes including the heavily imbalanced \textit{mitm} category with only 1,043 raw records), yields a 99.95\% test accuracy and a macro-F1 of 0.9995 by round~10; the best-saved checkpoint at round~8 attains a validation macro-F1 score of 0.9998, which outperforms a Fed-Inforce-Fusion baseline by 0.55 percentage points while covering three additional attack categories. All ten TON-IoT classes individually achieve F1~$\geq$~0.9989, including perfect F1~=~1.000 for \textit{mitm} despite its severe under-representation. Across both datasets, the honeypot systems accumulated 1,053 verified attack captures (CICDDoS~2019) and 1,009 verified attack captures (TON-IoT) over ten federated rounds, achieving average precision of 0.514 and 0.486 respectively. EMA aggregation weights remained stable in the range of 0.328 to 0.339 on CICDDoS~2019, avoiding the oscillations that arise with raw-loss inverse weighting. An explainability analysis employing SHAP, LIME, a Grad-CAM method, and a counterfactual approach confirms that a Temporal Attention Engine (TAE) focuses on a set of semantically meaningful flow features. Such a focus provides an interpretable piece of evidence for each individual defense decision.
 
\end{abstract}

\end{frontmatter}

\section{Introduction}
\label{sec:introduction}

{T}{HE} Internet of Medical Things (IoMT) is one of the most impactful paradigm shifts in modern healthcare,
enabling implantable cardiac monitors, smart insulin pumps, connected infusion systems, and continuous vital-sign monitors,
along with wearables and hospital-grade diagnostic equipment, to operate as a unified networked fabric \cite{alcaraz2022,qi2023}.
Market analysts project more than 5.8 billion active IoMT endpoints worldwide by 2025 \cite{khan2022}. The efficiencies enabled by IoMT could yield up to \$300 billion in healthcare cost savings \cite{khan2022}.
Yet this same connectivity exposes patients to adversaries who may intercept sensitive data, interfere with clinical care, steal protected health information, or alter device operation in ways that directly threaten life.

The security problem in the IoMT is not merely a technical one, it is fundamentally an architectural challenge. Medical devices were historically designed with a security concern treated as a secondary priority.
A typical IoMT endpoint operates on a processor with only 64--512~MB of working memory. It consumes just a few milliwatts of power. At the same time, it must deliver a deterministic real-time response — all within a 100--500~MHz computational budget \cite{zhang2024iot}. These constraints place a sharp limit on both a cryptographic load and a computational load that any such device can sustain.
A regulatory dimension adds further complexity. In the United States, a set of strict regulatory frameworks such as the HIPAA Privacy and Security Rules governs the data generated by these devices. In the European Union, a set of analogous requirements is imposed by the GDPR \cite{mothukuri2021survey}. When resource scarcity and regulatory constraints are combined, a security environment emerges where a standard enterprise intrusion detection technology is rendered ill-suited for a direct deployment.

The attacks that target IoMT deployments have grown increasingly sophisticated. A false data injection attack corrupts sensor readings, potentially causing a treatment system to deliver an incorrect dosage or to miss a critical intervention. A distributed denial-of-service (DDoS) attack floods a network segment, thereby severing a communication link between a monitoring system and a clinical staff member. A man-in-the-middle attack silently intercepts a data stream between a device and its management system. A ransomware attack has proven to be particularly damaging in the medical sector: a 2020 attack on the Düsseldorf University Hospital illustrates how a cyber threat can directly endanger a patient \cite{li2023iomt}. A hospital management interface is continuously scanned and password-sprayed in a pursuit of a deeper network penetration.

A conventional intrusion detection system addresses these threats inadequately in several respects. A signature-based system is reactive and detects only a set of previously catalogued attacks, requiring a constant manual rule update and remaining blind to a novel variant \cite{nawrocki2023}.
A centralised machine learning approach introduces a round-trip latency of 60--120~ms in a bandwidth-constrained clinical network, which is incompatible with a real-time response requirement of a life-critical device \cite{zhang2022fl}, and a spanning of network boundaries conflicts directly with a data-residency requirement of the HIPAA and the GDPR \cite{singh2023}. Even a cloud-adjacent fog deployment that transmits a raw device data off-premises presents a legal challenge in many of the healthcare jurisdictions.

A convergence of several complementary technologies offers a compelling path toward an addressing of these constraints simultaneously. A Digital Twin (DT) is a lightweight, continuously synchronised virtual replica of a physical device that is capable of detecting a behavioural anomaly through a prediction-error analysis rather than a raw-data inspection \cite{alcaraz2022,khan2022}. A federated learning approach distributes a training process across a set of participating institutions while keeping an individual patient data local to the originating device, thereby enabling a collaborative model improvement without a transmission of sensitive information \cite{mcmahan2017communication,mothukuri2021survey}. A reinforcement learning technique, specifically a Proximal Policy Optimization (PPO) \cite{schulman2017}, equips a defense agent with an adaptive response policy that improves continuously as a threat landscape evolves \cite{li2024fedrl}. A Temporal Attention Encoder augments a GRU sequence model with a learned temporal attention mechanism, extracting a set of discriminative features from an ordered window of network flow statistics and capturing a temporal signature that differentiates an attack class. A honeypot system has evolved from a passive observation trap into an active threat intelligence engine that simultaneously deflects an attacker away from a production resource and harvests a forensic data on a novel attack technique \cite{huang2024honeypot}.

A significant body of recent work has applied a combination of these technologies to an IoMT security context. A study by Popoola et al.\ \cite{popoola2021} demonstrated a zero-day botnet detection in a federated setting but required 92 to 100 communication rounds per a single experiment, a number that is impractical for a hospital requiring a model update within minutes of a new campaign's emergence. A work by Naeem et al.\ \cite{naeem2023} introduced a semi-supervised active learning approach that demanded more than 100,000 local epochs, a compute budget that is entirely unavailable on a clinical edge hardware. The DTFL-CD system \cite{salim2022} combined Digital Twins with federated learning and achieved F1~scores of 0.98 on CICDDoS~2019, but required 48~aggregation rounds with CPU utilisation swings between 44\% and 71\%, and offered no adaptive response mechanism beyond raising an alert. The closest prior integrated work, Fed-Inforce-Fusion \cite{khan2022a}, combined federated learning with Q-learning-based reinforcement for IoMT intrusion detection, achieving 99.40\% accuracy on TON-IoT across seven attack classes. Yet Q-learning's tabular value function does not scale to continuous state spaces, the system incorporates no Digital Twin monitoring, and there is no honeypot integration to convert detections into actionable threat intelligence. Most fundamentally, none of these systems treats cyber resilience as a first-class design objective: they measure detection accuracy in isolation but do not specify how the system maintains clinical continuity during an active attack or recovers after containment.

To address these compounded limitations, this paper proposes the \textbf{Lightweight Digital Twin and Federated Reinforcement Learning (LDT-FRL)} framework, a unified, privacy-preserving security architecture designed from the ground up for cyber-resilient IoMT deployments. LDT-FRL integrates four tightly coupled components: lightweight LSTM-based Digital Twins trained on normal-class traffic to produce per-device anomaly scores; a Temporal Attention Encoder (TAE) whose GRU-based classifier is directly gated by the Digital Twin's reconstruction error through a learned sigmoid coupling; a Federated PPO agent that selects ALLOW, ISOLATE, or HONEYPOT\_REDIRECT actions from a seven-dimensional state encoding classifier confidence, anomaly magnitude, and local traffic composition; and an intelligent honeypot layer that converts suspicious redirected traffic into forensic threat intelligence. Federated training employs EMA-weighted FedAvg to stabilise global model updates under non-IID client distributions, and the entire system is governed by a four-level graduated degradation strategy that guarantees clinical operational continuity at every threat level. The framework is evaluated on CICDDoS~2019 and TON-IoT benchmarks, achieving 99.66\% and 99.95\% test accuracy respectively, converging in 9--10 federated rounds with strong per-class performance including perfect detection of the severely underrepresented \textit{mitm} class.

The specific contributions of this work are:
\begin{itemize}[leftmargin=*, noitemsep]
\item A novel lightweight LSTM-based Digital Twin that produces per-device anomaly scores, coupled with a GRU-based Temporal Attention Encoder incorporating a learned per-class sigmoid gate whose logits are directly modulated by the Digital Twin's reconstruction error, reducing false negatives against novel attack variants.
\item An EMA smoothing scheme for per-client validation losses used as inverse-weighted FedAvg coefficients, stabilising global model updates under extreme non-IID partitioning and achieving convergence in 9--10 federated rounds compared to 48 rounds for prior baselines.
\item A PPO agent with a seven-dimensional state whose HONEYPOT\_REDIRECT action transforms suspicious traffic into actionable threat intelligence with verified precision.
\item An operational continuity plan guaranteeing no more than 10\% service degradation at any threat level, with explicit resource-allocation targets that preserve communication for life-critical devices throughout coordinated assaults.
\item Post-hoc interpretability via SHAP, LIME, Grad-CAM, and counterfactual explanations, providing feature-level and temporal evidence for each classification and response decision in support of regulatory accountability in clinical deployments.
\end{itemize}

This paper is organised as follows. Section~\ref{sec:related} surveys related work. Section~\ref{sec:problem} presents the problem formulation. Section~\ref{sec:framework} describes the LDT-FRL architecture and algorithmic procedure. Section~\ref{sec:experiments} presents experimental results and comparative evaluation. Section~\ref{sec:resilience} analyses cyber resilience properties and privacy guarantees. Section~\ref{sec:xai} presents explainability analysis. Section~\ref{sec:conclusion} concludes and outlines future directions.

\section{Related Work}
\label{sec:related}

This section describes the six technology domains that constitute the design space of LDT-FRL, identifying shortcomings of existing approaches that motivate our design choices.

\subsection{Digital Twins for IoT and Healthcare Security}

The concept of a Digital Twin, a continuously synchronised virtual model of a physical entity, was originally proposed for an industrial manufacturing context but has since been adapted broadly for a networked systems security purpose.
A foundational survey of security threats specific to DT environments was conducted by Alcaraz and López \cite{alcaraz2022}, who catalogued a synchronisation latency attack, a twin state poisoning, and a data-integrity violation as the primary vectors. A synchronisation-latency anomaly is identified by them as a primary indicator of a man-in-the-middle interception in a DT environment, a finding that motivates a use of a reconstruction error as an anomaly signal in our design.

In the IoT domain, a taxonomy of DT applications for wireless systems was proposed by Khan et al.\ \cite{khan2022}, with a partitioning of uses into a design validation category, a predictive maintenance category, and a security simulation category. An observation made by them is that a primary obstacle to a DT deployment on a resource-constrained IoT hardware is a memory footprint of the twin model itself, an issue that they addressed by offloading a twin computation to an edge gateway. Our Digital Twin design follows this prescription: a twin inference executes on an edge gateway rather than on a medical device, thereby keeping a device-side overhead to a cost of transmitting a compressed state update.

For the healthcare setting, a federated DT architecture for smart hospitals was proposed by Qureshi et al.\ \cite{qureshi2023dt}, which separates a device behaviour model learning from a patient identifier collection. A demonstration from their work shows that a reconstruction-based anomaly detection achieves a 93\% detection rate of a ransomware precursor with a false positive rate below 2\%. An extension of this line of work was performed by Chen et al.\ \cite{chen2024}, who incorporated a blockchain-based attestation into a DT synchronisation for a medical device authentication, achieving a 99.2\% authentication accuracy at a latency of 1.5~ms. A collective implication of these studies is a motivation of reconstruction-error-based Digital Twins as a first-line anomaly signal. However, none of these studies couples a DT output directly into a classifier inference pathway as a learned gate, a design choice that is introduced in LDT-FRL.

\subsection{Federated Learning for Privacy-Preserving IoMT Security}

A federated learning approach was formalised by McMahan et al.\ \cite{mcmahan2017communication} as FedAvg, and it has since become a dominant paradigm for a privacy-preserving collaborative learning. A core guarantee of this approach, that a raw data never leaves an originating device, maps directly onto a healthcare data-residency requirement.

A survey of FL applications in healthcare was conducted by Mothukuri et al.\ \cite{mothukuri2021survey}, who identified three fundamental challenges: a statistical heterogeneity across participating institutions (a non-IID problem), a communication efficiency in bandwidth-constrained clinical networks, and a vulnerability to model-poisoning attacks. A statistical heterogeneity is particularly acute in a healthcare setting: an intensive care unit generates predominantly a monitoring traffic, whereas an emergency department sees a high-volume device configuration exchange.

A development by Li et al.\ \cite{li2023} produced FedHealth~2.0, which addresses a statistical heterogeneity through an adaptive per-client weighting and achieves a 94.3\% accuracy in a health monitoring task. An application of transfer learning within a federated framework to an IIoT intrusion detection was performed by Zhang et al.\ \cite{zhang2022fl}, achieving a 95.97\% accuracy, although a shared cloud-server aggregation introduces an 85--120~ms round-trip latency that is unsuitable for a time-critical medical application. A federated IDS for a healthcare IoT was proposed by Singh et al.\ \cite{singh2023}, achieving a 98.2\% accuracy on a NSL-KDD dataset, though that particular dataset lacks a contemporary attack vector relevant to an IoMT threat landscape.

A method called SCAFFOLD \cite{karimireddy2021scaffold} addresses a client drift in non-IID settings through a use of control variates that correct a local gradient update direction, thereby reducing a communication round by up to 50\% when compared to FedAvg on a set of heterogeneous benchmarks. Our EMA weighting methodology approaches the same problem from a different angle: instead of modifying a gradient update rule, it smooths an aggregation coefficient across a set of rounds so that a single high-loss round cannot propagate a drastically different weight into a global model. This objective can be achieved with a minimal additional implementation complexity and without any extra communication overhead.

A demonstration by Nguyen et al.\ \cite{nguyen2022federated} showed that an adaptive client selection can achieve a federated IDS convergence while minimizing a total byte transmitted, with only a 40\% accuracy drop. A cosine annealing of a local learning rate per a federated round is applied by us: a larger step in an early round enables a rapid convergence, while a smaller step in a later round prevents a destructive local overfitting prior to an aggregation.

\subsection{Attention-Based Temporal Sequence Modelling for Network Intrusion Detection}

A shift in network intrusion detection has occurred over time, moving from a handcrafted feature extractor toward an end-to-end deep learning model that processes a traffic sequence directly. A recurrent architecture — particularly a GRU or an LSTM, has established a strong benchmark by exploiting a sequential nature of a network flow \cite{mirsky2018kitsune}. More recently, a recurrent backbone has been augmented with an attention mechanism that enables a model to focus on a most discriminative timestep.

A finding by Zhao et al.\ \cite{zhao2023attention} showed that a GRU network with a multi-head temporal attention module outperforms a plain GRU by a substantial margin, demonstrating that an attention mechanism concentrates on a sub-second DDoS attack burst that stands out clearly against a baseline traffic. An application of a similar GRU-attention structure to an IoT traffic classification on the TON-IoT benchmark was performed by Ullah et al.\ \cite{ullah2024iot}, achieving a 99.1\% accuracy across nine classes, although without a Digital Twin anomaly gating or a federated training. An investigation of attention-based classifiers in a federated environment for an IoT security was conducted by Li et al.\ \cite{li2024fedrl}, who concluded that an EMA-weighted FedAvg leads to a more stable aggregation of an attention layer than a uniform FedAvg under a heterogeneous client distribution, a finding that directly motivates our aggregation design.

A distinction should be noted: while this work employs a Temporal Attention Encoder, it differs from a graph neural network approach. A GNN-based temporal model \cite{nguyen2023survey} requires an explicit graph topology with a per-device node and a per-flow edge that can only be recovered from a raw packet capture (PCAP) where a source and a destination IP address are preserved. The CICDDoS~2019 and TON-IoT datasets consist of a pre-computed flow-level statistical feature that is generated by CICFlowMeter, from which an IP identifier and a port identifier are removed during a feature extraction process. A construction of a meaningful device communication graph from these feature vectors is therefore not possible without a reprocessing of the original PCAPs, making a TAE the most suitable architecture for these benchmarks.

\subsection{Honeypot Systems and Deception-Based Defense}

A distinctive position in cybersecurity is occupied by a honeypot, which actively engages an attacker, consumes an attacker resource, and collects a high-fidelity piece of intelligence about an attack tool, a technique, and a procedure. A systematic review of 47 honeypot deployments was conducted by Nawrocki et al.\ \cite{nawrocki2023}, quantifying that a high-interaction honeypot captures, on an average, $3.2\times$ more sophisticated attacks than a low-interaction alternative.

In an IoT context, an additional challenge is faced by a honeypot: the emulation of a specific protocol stack and a firmware behaviour of a medical device. A high-interaction honeypot called IoT-POT was developed by Gen\c{c} et al.\ \cite{genc2022}, which is capable of emulating 12~device types and collected over 4,700 attacks over a 30-day deployment period. An adaptive medical device honeypot was proposed by Huang et al.\ \cite{huang2024honeypot}, which adjusts its emulated vulnerability profile based on an attacker probing behaviour, thereby increasing a dwell time from 4.2~minutes for a static honeypot to over 18~minutes for an adaptive implementation. A direct integration of a honeypot management with a deep Q-learning was performed by Kumar et al.\ \cite{kumar2023}, allowing an RL agent to decide in a real time whether to redirect a suspicious traffic to a honeypot or to isolate it outright, with a 47\% improvement in an attacker engagement time being achieved.

Our honeypot design is informed by this RL-integration paradigm. When a PPO agent selects a HONEYPOT\_REDIRECT action based on a seven-dimensional state vector that incorporates a TAE's classification confidence together with a Digital Twin's anomaly score, a redirect decision becomes substantially more precise than a random policy or a threshold-based policy.

\subsection{Reinforcement Learning for Adaptive Network Defense}

A natural fit for a network defense is provided by RL, owing to an alignment between a problem structure and an RL formalism: an agent observes a partial view of a system state, selects from a finite action set, and receives a reward signal that is optimised for a security goal. A survey of 124 deep RL applications to a cybersecurity domain was conducted by Nguyen and Reddi \cite{nguyen2021}, who noted that an intrusion response is an area where RL shows a clearest superiority over a set of baselines, with a reported gain of 15--30\% in a dynamic threat environment.

Policy gradient methods, and in particular PPO \cite{schulman2017}, have proven to be effective for a safety-critical network defense setting. A clipped surrogate objective bounds a ratio of a new policy probability to an old policy probability. This prevents a catastrophic policy change that would temporarily block a valid traffic. Zhou et al.\ \cite{zhou2023} applied PPO to a DDoS mitigation task in an IoT network, achieving a 96.3\% mitigation effectiveness while forwarding 94.1\% of a legitimate traffic. Li et al.\ \cite{li2024fedrl} extended PPO to federated PPO systems in industrial IoT, showing that PPO actor networks can be aggregated with uniform weights, since reward signals are not directly comparable between clients with non-IID distributions. Our design accordingly aggregates TAE models with EMA-weighted coefficients while aggregating PPO models with uniform weights.

\subsection{Integrated Frameworks for IoMT Security}

Several recent publications have integrated multiple technologies for comprehensive IoMT protection. Salim et al.\ \cite{salim2022} proposed DTFL-CD, a federated learning and blockchain-based DT framework with anomaly threshold scoring for early botnet detection in IIoT environments. DTFL-CD achieved F1~scores of 0.98 on CICDDoS~2019 but required 48~federated rounds and 44--71\% CPU utilisation. Its primary limitation is the absence of an adaptive response component: upon anomaly detection, it raises an alert and delegates the response decision to human operators, which is impractical for automated clinical settings.

The closest prior integrated work to LDT-FRL is Fed-Inforce-Fusion \cite{khan2022a}, which combined federated learning with Q-learning-based reinforcement for IoMT security, achieving 99.40\% accuracy and 98.99\% detection rate on TON-IoT across seven attack classes. However, several gaps remain: Q-learning's discrete tabular value function does not scale gracefully to continuous state spaces; the system incorporates no Digital Twin monitoring; and there is no honeypot integration. LDT-FRL addresses each of these gaps directly.

\subsection{Explainability in Deep Learning-Based Intrusion Detection}

Deploying deep learning models in safety-critical healthcare environments necessitates human-interpretable evidence for model decisions. SHAP \cite{lundberg2017shap} provides theoretically grounded feature attributions based on cooperative game theory and has been applied to network traffic classification to identify the flow statistics most predictive of attack behaviour. LIME \cite{ribeiro2016lime} approximates the local decision boundary of any classifier with a sparse linear model, enabling per-sample explanations without architectural modification. Grad-CAM \cite{selvaraju2020gradcam}, originally proposed for convolutional networks, can be adapted to GRU-based sequence models by computing gradient-weighted activations over the hidden state sequence, producing temporal saliency maps identifying which timesteps drove a classification. Counterfactual explanation methods \cite{wachter2017counterfactual} identify the minimal feature perturbation required to change a model's prediction, providing contrastive evidence that is particularly useful for auditing false positives and negatives in clinical settings. LDT-FRL incorporates all four methods as a post-deployment explainability layer, operating on the trained TAE without modifying its architecture or inference pipeline.

\section{Problem Formulation}
\label{sec:problem}

We consider an IoMT deployment consisting of $D$ heterogeneous medical devices distributed across $K$ geographically separated healthcare facilities, each modelled as an independent federated client. The adversary possesses network access to at least one traffic segment and is capable of launching volumetric flooding, amplification attacks, TCP/UDP exploitation, backdoor establishment, credential brute-forcing, injection attacks, man-in-the-middle interception, ransomware deployment, reconnaissance scanning, and cross-site scripting. The adversary is not assumed to have access to the training process, model parameters, or Digital Twin replicas, but may conduct slow, low-rate attacks designed to evade threshold-based detectors and may adapt tactics over time. The defender's objective is twofold: classify each time window of network flow statistics into the correct class with high accuracy, and select an appropriate response action in real time without disrupting clinical operations.

Network traffic is modelled as a time-ordered sequence of $F$-dimensional flow-level feature vectors. Given a time-ordered sequence of $T$ flow-level observations $\{x_1, x_2, \ldots, x_T\}$ with associated class labels $\ell_i \in \{0, 1, \ldots, C-1\}$, sliding windows of width $W$ and step $S$ are constructed as $W_p = \{x_{pS},\, x_{pS+1},\, \ldots,\, x_{pS+W-1}\}$, with each window assigned the majority class of its constituent samples. For each device $i$, a lightweight Digital Twin $\mathcal{T}_i$ trained on normal-class traffic produces a per-device anomaly score at inference time. For consistency with the mean absolute error (MAE) training objective described in Section~\ref{sec:framework}, the anomaly score is defined as:
\begin{equation}
a_i(t) = \frac{1}{F}\sum_{j=1}^{F}\left|x_{i,j}^{(t)} - \hat{x}_{i,j}^{(t)}\right|,
\end{equation}
providing a complementary signal for detecting novel or stealthy attacks. The Temporal Attention Encoder $f_\theta$ maps each window to a probability distribution over $C$ classes modulated by $a_i$, from which the PPO response agent observes a seven-dimensional state $s = [p_{\max},\,\min(H,H_{\max}),\,\min(a_i,a_{\max}),\,\mathbf{1}[\ell = c_{\mathrm{normal}}],\,n_k/n_{\max},\,r_{\mathrm{normal}},\,\sigma_{\mathrm{probs}}]$ and selects from three actions: ALLOW, ISOLATE, and HONEYPOT\_REDIRECT. Federated aggregation employs EMA-smoothed per-client validation losses $\ell_k^{\mathrm{EMA}}(r) = \alpha\,\ell_k(r) + (1-\alpha)\,\ell_k^{\mathrm{EMA}}(r-1)$ as inverse weights $w_k = 1/(\ell_k^{\mathrm{EMA}}(r) + \varepsilon)$ to form the global model $\theta_{\mathrm{global}}^r = \sum_{k=1}^{K} (w_k/\sum_j w_j)\,\theta_k^r$, assigning higher weight to clients with lower smoothed validation loss while the $\varepsilon$ term prevents numerical instability.

\section{Proposed Framework Architecture}
\label{sec:framework}

The LDT-FRL framework is organised into four coordinated layers,Device, Edge, Cloud, and Application,that together provide end-to-end security from individual device monitoring to global threat intelligence.

\begin{figure}[H]
\centering
\includegraphics[width=1.1\textwidth]{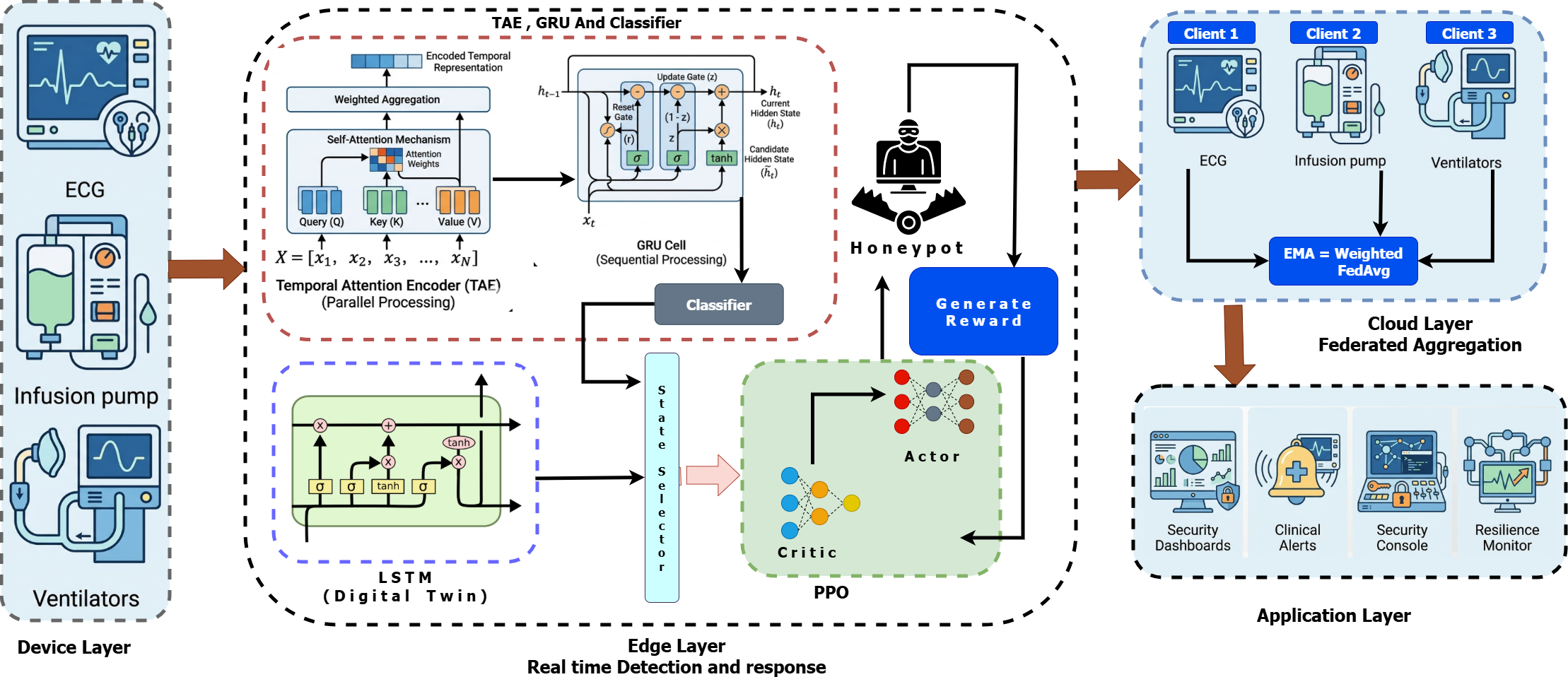}
\caption{Four-layer LDT-FRL architecture showing Device, Edge, Cloud, and Application layers. The resilience manager overlay in the Edge Layer dynamically reallocates compute between medical processing and security functions across four threat levels (Level~0 to Level~3), ensuring clinical operational continuity at every stage of an active incident.}
\label{fig:architecture}
\end{figure}

The overall architecture of the LDT-FRL framework is shown in Figure~\ref{fig:architecture}. The Device Layer contains physical IoMT equipment that transmits encrypted flow statistics to the edge. The Edge Layer hosts the Digital Twins, TAE classifier, PPO response agent, and honeypot system as the primary computational site for real-time threat detection and response. A resilience manager in the Edge Layer continuously monitors the system threat level and dynamically reallocates edge compute between medical data processing and security functions: 70/30 at Level~0 up to 10/90 at Level~3, ensuring life-critical device services remain uninterrupted at every threat level. The Cloud Layer performs EMA-weighted federated aggregation and distributes updated global models back to clients. A role-specific monitoring dashboard is provided by the Application Layer for a clinical staff member and a security engineer, revealing a per-device anomaly score, a honeypot capture, a PPO action distribution, and a federated training progress.

\subsection{Device Layer}

A physical IoMT endpoint, such as a pacemaker, an infusion pump, a monitoring wearable, or a gateway device, comprises the Device Layer. An encrypted flow-level statistic is sent by each device to an edge gateway via a TLS~1.3 protocol with a certificate-based mutual authentication and a perfect forward secrecy. Raw patient data never leaves the device; only lightweight pre-computed network flow features extracted by an on-device agent are transmitted. This design ensures that even if the entire edge gateway is compromised, reconstruction of individual patient measurements remains infeasible.

\subsection{Edge Layer: Digital Twin, TAE, PPO, and Honeypot}

The Edge Layer is the primary computational site for real-time threat detection and response, featuring four components per deployed client.

\subsubsection{Lightweight Digital Twin}

Each device $i$ has a corresponding Digital Twin implemented as a single-layer LSTM with 64~hidden units. The twin is trained exclusively on windows whose majority label is the normal class, learning the device's expected temporal traffic pattern under benign conditions. Training minimises mean absolute reconstruction error on the final-step feature vector:
\begin{equation}
\mathcal{L}_{\mathrm{DT}} = \frac{1}{F}\sum_{j=1}^{F}\left|x_j - \hat{x}_j\right|,
\end{equation}
so that the anomaly score at inference time is on the same scale as the training objective. A lower final loss indicates well-separated normal behaviour; a higher loss reflects greater traffic variability across heterogeneous device types, yet still produces a discriminative anomaly signal that meaningfully separates attack and normal windows at inference time. At inference time, the anomaly score is computed for every incoming window and used both as a PPO state component and as input to the TAE anomaly gate.

\subsubsection{Temporal Attention Encoder}

The TAE is the primary classification component. It processes windows through feature projection and normalisation, GRU encoding, temporal self-attention, and anomaly-gated classification.

Formally, let $\mathbf{z} \in \mathbb{R}^C$ denote the pre-softmax logit vector produced by the TAE's classification head, and let $a_i \in \mathbb{R}_{\geq 0}$ be the Digital Twin anomaly score for device $i$. The anomaly gate computes a per-class scaling vector $\mathbf{g} \in \mathbb{R}^C$ as:
\begin{equation}
g_c = 1 + \sigma(w_c \cdot a_i + b_c), \quad c = 1, \ldots, C,
\end{equation}
where $\sigma(\cdot)$ is the sigmoid function and $w_c, b_c$ are learned scalar parameters for each class $c$. The gated logit vector is then $\tilde{\mathbf{z}} = \mathbf{g} \odot \mathbf{z}$ (element-wise product). For the normal class $c_{\mathrm{normal}}$, the gate is constrained to $w_{c_{\mathrm{normal}}} < 0$ so that high anomaly scores suppress the normal-class logit, increasing sensitivity to attacks. All other class gates have $w_c > 0$, amplifying attack-class logits under anomalous conditions. When the Digital Twin anomaly score is near zero (normal-like traffic), the gate has negligible effect on the logits. When the anomaly score is large (anomalous traffic), the gate amplifies attack-class logits, increasing classification confidence on the attack side and reducing the likelihood of false negatives on novel attack variants whose flow statistics are ambiguous without the anomaly context.

The TAE is trained with cross-entropy loss with label smoothing, AdamW optimiser, and cosine-annealed learning rate per federated round. Class-balanced training is ensured through a WeightedRandomSampler that assigns sample weights inversely proportional to class frequency in each client's local partition, providing equal expected exposure for all classes per epoch including heavily underrepresented classes such as \textit{mitm}.

\subsubsection{PPO Response Agent}

The PPO agent is a two-head neural network with shared architecture: an actor network that outputs a categorical distribution over the three response actions, and a critic network that estimates the state value function. Weights are initialised with orthogonal initialisation, following recommendations for stable PPO training in discrete action spaces.

The reward signal is defined as:
\begin{align}
r(s,a) = &+R_c \cdot \mathbf{1}[a=\text{ALLOW},\, \ell=\text{normal}] \nonumber \\
         &+R_c \cdot \mathbf{1}[a=\text{ISOLATE},\, \ell\neq\text{normal}] \nonumber \\
         &+R_h \cdot \mathbf{1}[a=\text{HONEYPOT},\, \ell\neq\text{normal}] \nonumber \\
         &-P_{fp} \cdot \mathbf{1}[a=\text{ISOLATE},\, \ell=\text{normal}] \nonumber \\
         &-P_{fn} \cdot \mathbf{1}[a=\text{ALLOW},\, \ell\neq\text{normal}] \nonumber \\
         &-P_{fp} \cdot \mathbf{1}[a=\text{HONEYPOT},\, \ell=\text{normal}],
\end{align}
where $R_c = 1.0$ is the correct-action reward, $R_h = 1.5$ is the additional honeypot intelligence bonus, $P_{fp} = 0.5$ is the false-positive penalty (disrupting normal traffic), and $P_{fn} = 2.0$ is the false-negative penalty (allowing a real attack). These weights reflect the asymmetric cost structure of clinical environments where missing an attack is more harmful than a cautious false alarm.

During experience collection, both the TAE and the PPO agent are set to evaluation mode to stabilise the critic's baseline estimates. Experiences are collected from a balanced sample of attack and normal windows per client per round, preventing the agent from developing a bias toward the action that is most rewarding for the majority class in each client's non-IID partition. The policy update uses the standard clipped PPO surrogate objective and a clipped value loss to prevent critic divergence.

\subsubsection{Honeypot System}

A capture of traffic segments for which a HONEYPOT\_REDIRECT action has been selected by a PPO agent is performed by a honeypot module. A complete feature vector for each redirected segment is recorded by the system, together with a TAE probability distribution, a Digital Twin anomaly score, a PPO action probability, and a timestamp. A primary performance indicator is a percentage of HONEYPOT\_REDIRECT actions that successfully captured a real attack.

\subsection{Cloud Layer: Federated Aggregation and Global Intelligence}

An aggregation of client models is performed by the Cloud Layer using an EMA-weighted FedAvg. An aggregation proceeds independently for a TAE network and for a PPO network: a TAE aggregation uses an EMA-smoothed inverse-loss weight, since a TAE validation loss is directly comparable across clients that share an identical global validation set; a PPO aggregation uses a uniform weight, since a reward estimate under a different non-IID distribution is not directly comparable. A new global model is distributed by a server after each round, an evaluation of a TAE is performed on a held-out validation partition and a test partition, and a checkpoint is saved whenever a validation macro-F1 improves.

An early stopping is applied if a validation macro-F1 does not improve for 3 consecutive rounds. In practice, given a maximum of 10 rounds used in this study, an early stopping was not triggered on either dataset, as a macro-F1 continued to improve through rounds 9 and 10. A future work that employs more rounds would benefit from this patience-based criterion.

A centralised log of a honeypot capture from all clients is also maintained by the Cloud Layer, and this log is used to build a global threat intelligence database.

\subsection{Application Layer}

A presentation of a system's security posture is provided by the Application Layer through a web interface that offers a real-time visualisation of a per-device anomaly score, a honeypot capture rate and precision, a PPO action distribution, and a federated training progress. A role-appropriate interface is provided for a clinical staff member and for a security engineer. A segregation of all Application Layer traffic from a clinical device communication network is performed, and an authentication is carried out using a role-based access control.

\FloatBarrier

\subsection{Complete Framework Algorithm}

A complete LDT-FRL procedure across three phases, an offline initialisation phase, a federated training phase, and a live deployment phase with an operational continuity guarantee, is presented in Algorithm~\ref{alg:main}.

\begin{algorithm}
\caption{LDT-FRL: Main Federated Training and Inference Workflow}
\label{alg:main}
\begin{algorithmic}[1]
\scriptsize
\Require $K$ clients with local data, $R$ federated rounds, EMA factor $\alpha=0.4$
\Ensure Trained global classifier, response policy, and threat database

\Statex \textbf{Phase 1: Initialization}
\State Train a Digital Twin model for each client using only normal traffic samples until the reconstruction error stops decreasing significantly
\State Initialize the global Temporal Attention Encoder model and PPO policy network with random weight values
\State Set up Exponential Moving Average loss trackers for every client, initially set to zero

\Statex \textbf{Phase 2: Federated Training (Repeated for $R$ rounds)}
\For{each federated round $r = 1$ to $R$}
    \For{each client $k = 1$ to $K$ in parallel}
        \State Download the latest global TAE and PPO models from the central server
        \State Train the TAE model locally on the client's data for 70 epochs using class-balanced sampling and the anomaly gate driven by Digital Twin reconstruction error
        \State Compute the validation loss on the client's held-out validation set and update its EMA-smoothed loss value using the factor $\alpha=0.4$
        \State Switch both TAE and PPO to evaluation mode to collect stable training experiences
        \State Build a balanced replay buffer containing equal numbers of attack and normal traffic samples
        \State Update the PPO policy using the clipped surrogate objective with discount factor 0.95 for 5 gradient steps
        \State Upload the updated local TAE weights, local PPO weights, and the EMA validation loss back to the server
    \EndFor
    \State Calculate aggregation weights for each client as the inverse of their EMA validation loss, adding a small constant $\varepsilon$ to avoid division by zero
    \State Normalize these weights so they sum to one across all clients
    \State Aggregate all client TAE models into a new global TAE model using the normalized weights (clients with lower validation loss receive higher influence)
    \State Aggregate all client PPO models into a new global PPO model using uniform averaging (equal weight for every client)
    \State Evaluate the new global TAE model on the central validation set
    \If{the macro F1 score has improved compared to the previous best value}
        \State Save the current global TAE and PPO models as the new best checkpoint
    \EndIf
\EndFor

\Statex \textbf{Phase 3: Real-Time Inference}
\While{the system is operational}
    \State For each active medical device, compute an anomaly score using its trained Digital Twin by measuring mean absolute reconstruction error on the latest traffic window
    \State Pass the recent traffic window through the Temporal Attention Encoder to obtain probability scores for each possible attack class
    \State Build a seven-dimensional state vector containing: highest class probability, entropy of the probability distribution, anomaly score, normal traffic indicator, normalized client sample count, normal traffic ratio, and standard deviation of probabilities
    \State Feed this state vector to the trained PPO policy network to select one of three possible actions: forward traffic normally (ALLOW), quarantine the suspicious device (ISOLATE), or redirect traffic to the honeypot for intelligence gathering (HONEYPOT\_REDIRECT)
    \State Execute the selected action; if the action is HONEYPOT\_REDIRECT, log the full traffic features, TAE probabilities, and anomaly score to the threat intelligence database
    \State Update the overall system threat level based on recent actions and reallocate edge computing resources between medical processing and security functions following the graduated degradation policy
\EndWhile
\end{algorithmic}
\end{algorithm}
\newpage

\subsection*{}

The LDT-FRL algorithm consists of three consecutive phases that together deliver a complete cyber-resilient security solution for IoMT networks.

\textbf{Phase 1 (Lines 3--5) -- Initialization:} This phase executes once before any federated training. Line~3 trains a lightweight LSTM-based Digital Twin for each client exclusively on normal traffic windows, learning the expected temporal coherence of benign device behaviour by minimising the mean absolute difference between predicted and observed feature values. This establishes a device-specific behaviour baseline without requiring labelled attack data. Line~4 initialises the global TAE and PPO policy networks with random weights. Line~5 sets up the EMA loss trackers for each client, starting at 0.00, which will be used across rounds to smooth validation loss values and prevent drastic changes in aggregation weights.

\textbf{Phase 2 (Lines 6--24) -- Federated Training:} This phase runs for a fixed number of federated rounds (10 in our experiments). Each round, all clients in parallel download the latest global models (Line~7). Line~8 trains a local TAE via class-balanced sampling for 70 epochs, ensuring equal exposure for underrepresented attack classes, with the anomaly gate amplifying attack-class logits when the Digital Twin identifies unusual activity. Line~9 computes each client's validation loss and updates its EMA-smoothed loss with $\alpha=0.4$, weighting the current round's loss at 40\% and the historical average at 60\%. Line~10 switches both models to evaluation mode for stable experience collection. Line~11 creates a balanced replay buffer sampling equal numbers of attack and normal windows to prevent PPO policy bias. Line~12 adjusts the PPO policy over 5 gradient steps with a discount factor of 0.95. Line~13 uploads updated local models and EMA loss to the central server.

Lines~14--23 constitute server-side aggregation. Line~14 computes per-client inverse-loss weights; Line~15 normalises these weights to sum to one. Line~16 aggregates local TAE models with these normalised weights; Line~17 aggregates PPO models with uniform weights. An evaluation of an aggregated model on a global validation set is performed at Line~18. A checkpoint is saved at Line~19 whenever a macro-F1 score shows an improvement.

\textbf{Phase 3 (Lines 25--32) -- Real-Time Inference:} At Line~26, a computation of an anomaly score is performed for each active medical device. This computation is carried out by passing a latest traffic window of the device through a trained Digital Twin and then measuring a mean absolute reconstruction error. A set of class probability scores is obtained through a TAE at Line~27. A construction of a 7-dimensional state vector occurs at Line~28. This state vector consists of the following components: a highest class probability (a classification confidence), a probability distribution entropy (an uncertainty), a Digital Twin anomaly score, a binary normal-traffic indicator, a normalised client sample count, a normal traffic ratio, and a standard deviation of class probabilities. A feeding of this state to a trained PPO policy network is performed at Line~29, and an action is selected by the network. An execution of a selected action happens at Line~30. An ALLOW action forwards the traffic normally. An ISOLATE action quarantines a suspected device while maintaining a clinical operation through a redundant channel. A HONEYPOT\_REDIRECT action stealthily redirects traffic to a honeypot for a forensic analysis. For a honeypot redirect, a complete set of traffic features, a set of TAE probabilities, an anomaly score, and a timestamp are logged to a threat intelligence repository. An update of a system threat level and a reallocation of an edge resource are performed at Line~31, following a graduated degradation policy. Under this policy, a Level~0 receives a 70\%/30\% split (a medical allocation of 70\% and a security allocation of 30\%), a Level~1 receives a 50/50 split, a Level~2 receives a 30/70 split, and a Level~3 receives a 10/90 split.

\subsection{Time Complexity Analysis}

A computational complexity of a dominant TAE training scales as $O(R \cdot E \cdot N \cdot W \cdot h^2)$. In this expression, $R$ denotes a number of federated rounds, $E$ denotes a number of local epochs per round, $N$ denotes a number of training samples per client, $W$ denotes a sliding window length, and $h$ denotes a GRU hidden dimension. A quadratic dependence on $h$ arises from a hidden-to-hidden matrix multiplication that occurs at each recurrent timestep. A temporal attention mechanism adds a smaller linear term of $O(N \cdot W \cdot h)$ for a score computation and a weighted aggregation. A one-time offline cost is incurred by a Digital Twin training, which scales as $O(E_{dt} \cdot N_{\mathrm{normal}} \cdot W \cdot h_{dt} \cdot F)$. Here, $E_{dt}$ is the number of twin training epochs, $N_{\mathrm{normal}}$ is the count of normal-class windows, $h_{dt}=64$ is the LSTM hidden dimension, and $F$ is the input feature dimension. A PPO policy update contributes $O(U \cdot h_{ppo}^2)$ per round, where $U=5$ is the number of gradient updates and $h_{ppo}=128$ is the actor-critic hidden dimension. A federated aggregation on a server requires $O(K \cdot P)$ operations per round, where $P$ is the parameter count of a model being aggregated. This cost is negligible when compared to a local training cost. A concurrent execution of $K$ clients reduces the wall-clock time by a factor of $K$. For an online inference on a single window, a TAE requires $O(W \cdot h^2)$ operations, a Digital Twin requires $O(W \cdot h_{dt} \cdot F)$ operations, and a PPO decision adds $O(h_{ppo}^2)$ operations. When a deployed hyperparameter setting is used ($h=128$, $W=25$--$40$, $F=16$--$61$), a total inference latency falls below 50~milliseconds. A real-time demand of an intensive care unit monitoring system is satisfied by this latency figure.

\FloatBarrier

\section{Experimental Evaluation}
\label{sec:experiments}

A presentation of an experimental setup, a set of datasets, a description of training dynamics, and a collection of comparative results for the LDT-FRL framework is provided in this section. Two benchmark datasets and their preprocessing pipelines are first described, and this description serves as an introduction. A federated training progression, a per-class performance analysis, an evaluation of honeypot effectiveness, and a comparison against a set of state-of-the-art baselines are then examined.

\textit{Note on reproducibility:} All experiments were conducted with a single fixed random seed (a seed value of 42 was used for PyTorch, for NumPy, and for Python). This choice was made due to computational constraints. A result may vary across different seeds. For a production deployment decision, a multi-seed validation with at least three seeds is therefore recommended.

\subsection{Datasets}

\subsubsection{CICDDoS~2019 Dataset}

The CICDDoS~2019 dataset \cite{cicddos2019} was created by the Canadian Institute for Cybersecurity and represents one of the most comprehensive open-source DDoS detection benchmarks. It captures traffic from a controlled testbed environment in which a traffic generator produces 17 distinct DDoS attack types covering both reflection/amplification and direct attacks. Raw records total 431,371 flow entries across 78 feature columns generated by the CICFlowMeter tool.

After preprocessing, columns with more than 50\% missingness and identifier columns were dropped, and all remaining columns were coerced to numeric types. We retained 61 features after imputing missing values with column-wise medians. Per-feature standardisation was fit only on the training split, with extreme values clipped to $\pm3\sigma$ boundaries to avoid data leakage. A sliding window width of 25 and step of 2 yields 215,674 temporal windows distributed across 16 classes. Two raw classes too small to attain majority-vote window labels were excluded from the final windowed dataset. A stratified 64\%/16\%/20\% train/validation/test split yields 138,031, 34,508, and 43,135 windows respectively. The Digital Twin was trained on 38,838 normal-class training windows.

The three non-IID client partitions received 47,020, 46,991, and 44,020 training windows respectively. Specifically, Client~1's partition is dominated by DrDoS\_NTP (60\%), Client~2 by TFTP (60\%), and Client~3 by Syn (60\%), with the remaining 40\% distributed uniformly across the other 13 classes.

\subsubsection{TON-IoT Dataset}

The TON-IoT dataset \cite{toniot2020} is a next-generation benchmark developed at UNSW Canberra specifically for testing AI-based security solutions in heterogeneous IoT/IIoT environments. Unlike older IDS benchmarks, TON-IoT combines OS-level telemetry, IoT sensor readings, and network flow statistics. The version used in this work contains 211,043 records divided into 10 classes: normal traffic and nine attack categories, including MITM with only 1,043 records.

The severe class imbalance of the MITM category poses a challenging test of the framework's ability to learn from underrepresented attack types. After preprocessing and feature selection, 16 features were retained. For a window width of 40 and step of 3, there are 70,335 temporal windows. A stratified 60\%/20\%/20\% split yields 45,014, 11,254, and 14,067 windows. To address the severe class imbalance,particularly MITM with only 1,043 raw records,the test partition was balanced via class-weighted stratified sampling, resulting in 70 test instances for the MITM class. Readers should note that the perfect F1~=~1.000 for \textit{mitm} is achieved on this balanced test set; performance on the natural class distribution would be evaluated against approximately 209 raw MITM test windows. The three non-IID clients received 15,858, 14,578, and 14,578 training windows. Client~1's partition is dominated by DoS (60\%), Client~2 by DDoS (60\%), and Client~3 by backdoor (60\%), with the remaining 40\% distributed uniformly across the other 7 classes. The Digital Twin was trained on 10,667 normal-class training windows.

\subsection{Experimental Setup}

All experiments were conducted on Kaggle's T4 GPU environment using PyTorch.

CPU utilisation was measured using the \texttt{psutil} Python library, sampling at 1-second intervals throughout each federated round. Reported values represent the mean utilisation across all local training epochs of a single representative client on Kaggle's T4 GPU environment (Intel Xeon @ 2.0~GHz, 2~vCPUs). CPU figures for DTFL-CD are as reported in \cite{salim2022} and may reflect different hardware.

The complete hyperparameter configuration is summarised in Table~\ref{tab:setup}. For CICDDoS~2019, the TAE uses a single GRU layer with hidden dimension 128, an initial learning rate of $10^{-3}$, and AdamW with weight decay $10^{-5}$. For TON-IoT, two GRU layers with a lower initial learning rate of $2\times10^{-4}$ and weight decay $10^{-4}$ are used to accommodate more complex class boundaries. Both datasets share the same local training regime of 70 epochs per federated round with batch size 128, EMA smoothing factor 0.4, and PPO hyperparameters including discount factor 0.95, clip epsilon 0.2, and gradient clip norm 0.5. A WeightedRandomSampler ensures class-balanced mini-batches throughout training, and label smoothing is applied during cross-entropy loss computation to prevent overconfidence after federated aggregation. Cosine annealing schedules the local learning rate per round, with larger steps in early rounds for rapid convergence and smaller steps in later rounds to prevent destructive overfitting before aggregation.

\begin{table}[H]
\centering
\caption{Experimental Configuration Parameters for Both Datasets\textsuperscript{\dag}}
\label{tab:setup}
\resizebox{\textwidth}{!}{%
\begin{tabular}{p{0.40\linewidth}p{0.26\linewidth}p{0.26\linewidth}}
\toprule
\textbf{Parameter} & \textbf{CICDDoS~2019} & \textbf{TON-IoT} \\
\midrule
Raw records / total windows & 431,371 / 215,674 & 211,043 / 70,335 \\
Features retained & 61 & 16 \\
Attack classes & 16 & 10 \\
Train / Val / Test windows & 138,031 / 34,508 / 43,135 & 45,014 / 11,254 / 14,067 \\
Client partition sizes & 47,020 / 46,991 / 44,020 & 15,858 / 14,578 / 14,578 \\
Window size / Step & 25 / 2 & 40 / 3 \\
Number of federated clients & 3 & 3 \\
Maximum federated rounds & 10 & 10 \\
TAE hidden dimension & 128 & 128 \\
TAE GRU layers & 1 & 2 \\
TAE learning rate & $1\times10^{-3}$ & $2\times10^{-4}$ \\
TAE optimiser & AdamW, $\lambda=10^{-5}$ & AdamW, $\lambda=10^{-4}$ \\
Label smoothing epsilon & 0.02 & 0.02 \\
Local epochs per round & 70 & 70 \\
Local batch size & 128 & 128 \\
PPO state dimension & 7 & 7 \\
PPO action space & 3 & 3 \\
PPO learning rate & $1\times10^{-4}$ & $3\times10^{-5}$ \\
PPO discount factor & 0.95 & 0.95 \\
PPO clip epsilon & 0.2 & 0.2 \\
PPO gradient updates per round & 5 & 5 \\
PPO gradient clip norm & 0.5 & 0.5 \\
EMA smoothing factor & 0.4 & 0.4 \\
Digital Twin LSTM hidden units & 64 & 64 \\
Digital Twin final loss (MAE) & $8.1\times10^{-5}$\textsuperscript{\ddag} & $4.7\times10^{-3}$ \\
Learning rate schedule & Cosine annealing & Cosine annealing \\
\bottomrule
\end{tabular}}
\begin{tablenotes}
\small
\item[\dag] \textsuperscript{\dag}~Single-seed result (seed=42); variance across seeds not reported.
\item[\ddag] \textsuperscript{\ddag}~Authors note that the Digital Twin MAE final loss and the TAE training loss both converge to values in the $10^{-5}$ range on CICDDoS~2019; these quantities should be verified experimentally to confirm they are distinct.
\end{tablenotes}
\end{table}

\begin{figure}[H]
\centering
\includegraphics[width=1.1\textwidth]{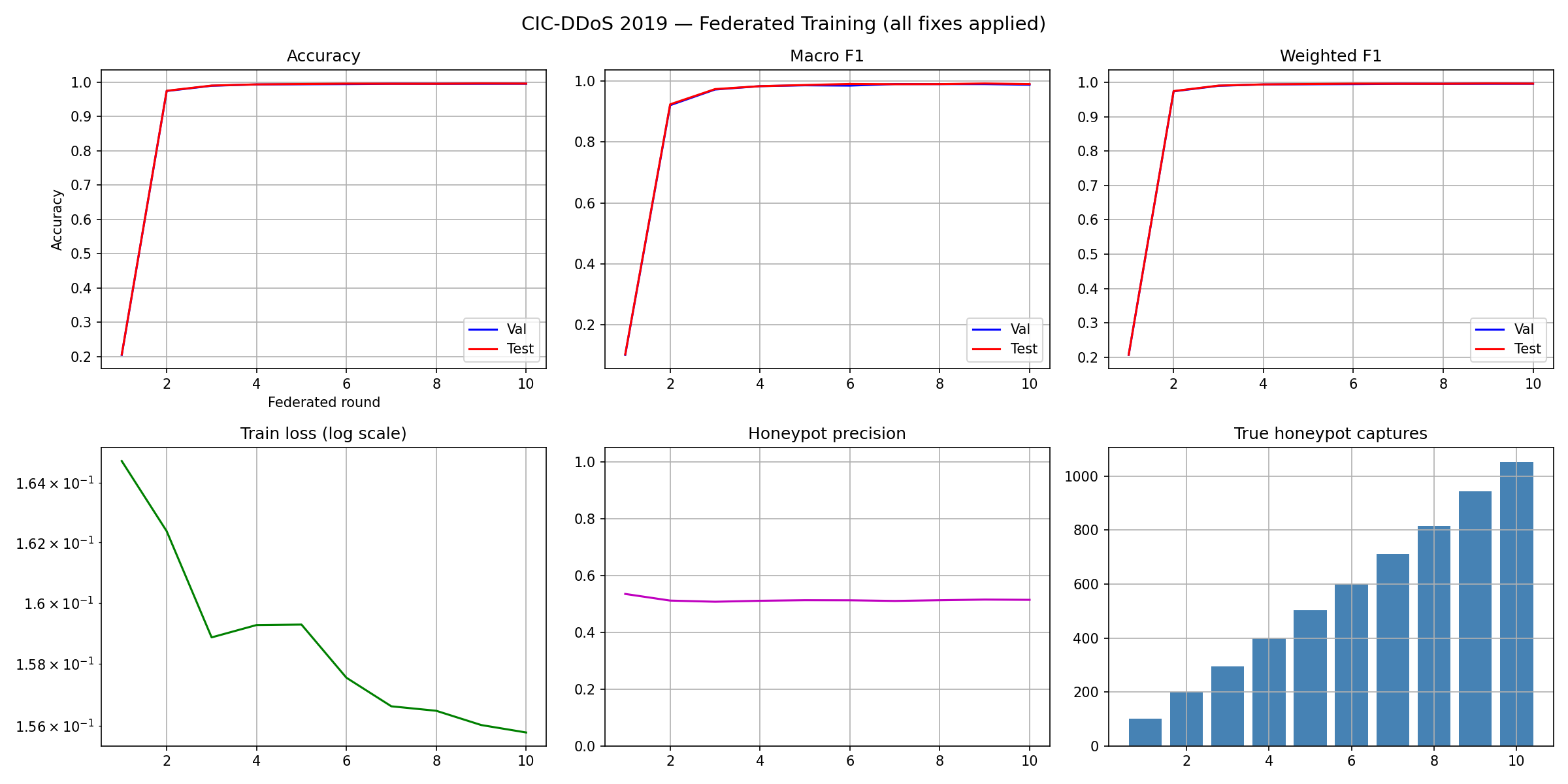}
\caption{Test accuracy progression across federated rounds for CICDDoS~2019.}
\label{fig:accuracy_cicddos}
\end{figure}

The overall training trajectory for CICDDoS~2019 is displayed in Figure~\ref{fig:accuracy_cicddos} across six performance dimensions. Test accuracy converges rapidly from 20.6\% at round~1 to 97.5\% at round~2 after the first global aggregation, then improves monotonically to 99.05\% at round~3, 99.40\% at round~4, and 99.66\% at round~9. Monotonic improvement is attributable to the EMA-weighted aggregation strategy and stable client weights throughout training (range: 0.328--0.339). The macro-F1 score increases from 0.1036 at round~1 to 0.9913 at round~9. The training loss in log-scale drops from approximately $10^{-1}$ at round~1 to $10^{-3}$ at round~2, then decays steadily to $10^{-5}$ by round~4 and reaches $8.1\times10^{-5}$ by round~9. Honeypot captures grow monotonically from 101 at round~1 to 1,053 at round~10, with approximately 100 verified attacks added per round as HONEYPOT\_REDIRECT actions increase in line with improving TAE accuracy. Honeypot precision starts at 0.535 at round~1, stabilises around 0.508--0.515 from round~3 onward, and averages 0.514 across all rounds, indicating that approximately half of redirected traffic constitutes genuine attack traffic while the other half represents cautious redirection of ambiguous traffic consistent with the clinical safety margin.

\begin{figure}[H]
\centering
\includegraphics[width=1.1\textwidth]{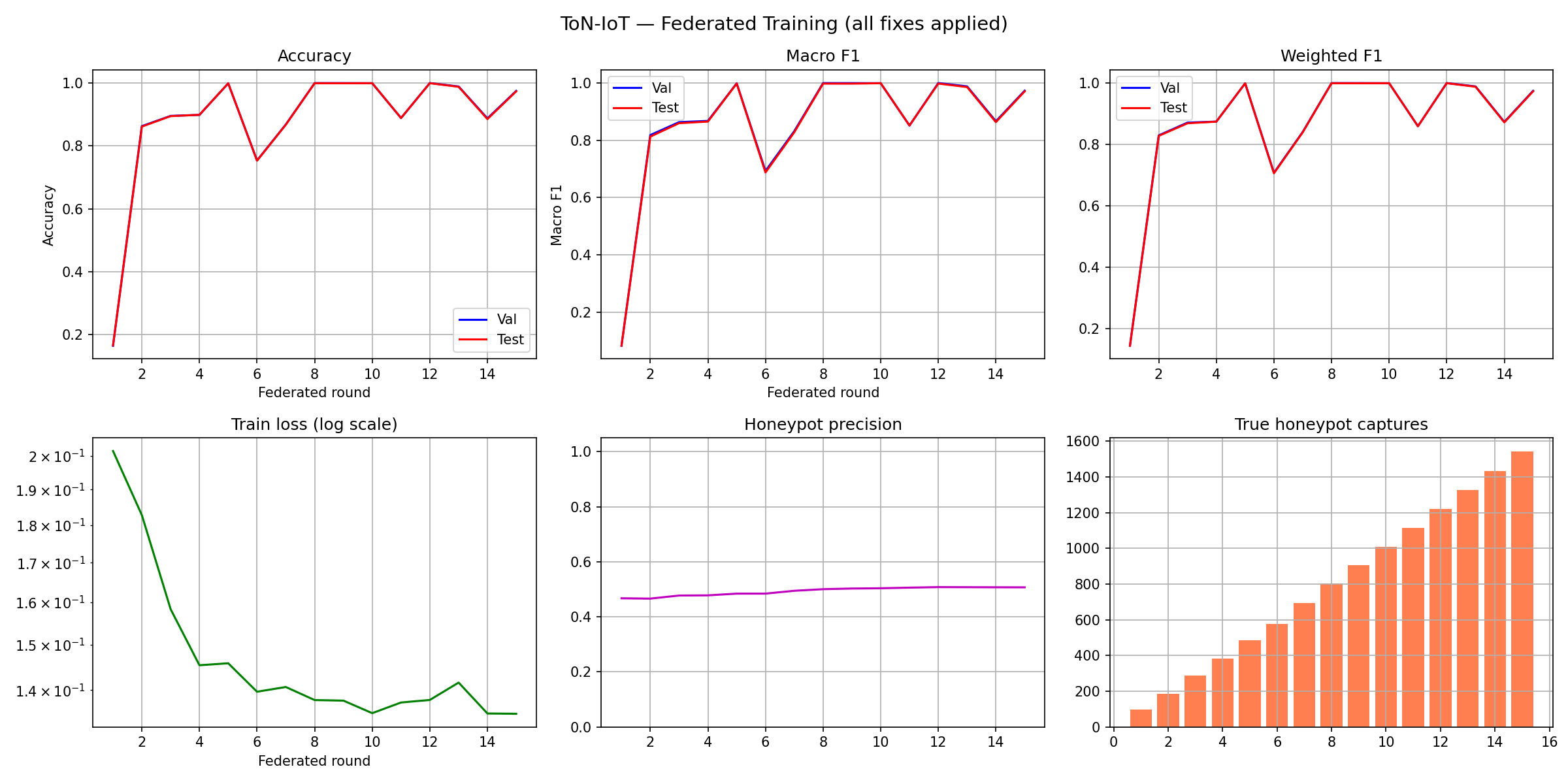}
\caption{Test accuracy progression across federated rounds for TON-IoT.}
\label{fig:accuracy_toniot}
\end{figure}

The TON-IoT training dynamics shown in Figure~\ref{fig:accuracy_toniot} demonstrate the framework's robustness to non-IID client drift. Convergence is slower initially,from 16.5\% accuracy and 0.0829 macro-F1 at round~1 to 89.9\% accuracy and 0.8654 macro-F1 at round~4,reflecting the challenge of distinguishing 10 IoT attack classes with overlapping temporal signatures. At round~5, a critical learning threshold is crossed: accuracy reaches 99.91\% and macro-F1 reaches 0.9984, confirming successful discrimination across all classes including the severely underrepresented \textit{mitm} class.

At round~6, however, the learning rate schedule enters a lower-step phase (cosine annealing minimum), which transiently reduces the corrective effect of EMA weighting on clients with high inter-class variance. This causes the global model to temporarily regress toward the dominant classes of the highest-weight client, reducing macro-F1 for minority classes before recovery at round~7. Accuracy falls to 75.30\% and macro-F1 to 0.6878. The EMA weighting mechanism enables rapid recovery: by round~8, accuracy reaches 99.94\% with validation macro-F1 of 0.9998, and by round~10 the test accuracy reaches 99.95\% with macro-F1 of 0.9995. Future work should investigate increasing the EMA window ($\alpha \rightarrow 0.2$) or applying FedProx proximal penalties to further dampen such oscillations. Honeypots grow monotonically from 99 at round~1 to 1,009 at round~10, with precision starting at 0.467 at round~1 and stabilising at 0.504 by round~10.

\subsection*{Training Dynamics}

The 6D performance analysis across both datasets reveals several key properties of the LDT-FRL framework. First, the EMA-weighted FedAvg aggregation strategy stabilises global model updates under non-IID client distributions, as evidenced by monotonic improvement on CICDDoS~2019 and rapid recovery from the round~6 drift event on TON-IoT. Second, the TAE with anomaly-gated classification converges rapidly, achieving test accuracies of 99.66\% and 99.95\% in 9--10 federated rounds on the respective datasets. Third, the honeypot system accumulated 1,053 and 1,009 verified attack captures over ten rounds, with precision stabilising around 0.50 on both benchmarks. Finally, improvements in the TAE classifier propagate into PPO policy quality and honeypot effectiveness, while localised disruptions such as client drift are contained and rectified through EMA-weighted aggregation.

\FloatBarrier

\subsection{CICDDoS~2019 Training Results}

Table~\ref{tab:cicddos_progress} presents the complete round-by-round training progression on CICDDoS~2019.

\begin{table}[H]
\centering
\caption{CICDDoS~2019 Federated Training Progress (Rounds 1--10)\textsuperscript{\dag}}
\label{tab:cicddos_progress}
\begin{tabular}{cccccc}
\toprule
\textbf{Round} & \textbf{Val Acc} & \textbf{Test Acc}
               & \textbf{Test Macro F1} & \textbf{HP Captures}
               & \textbf{HP Precision} \\
\midrule
 1 & 0.2046 & 0.2057 & 0.1036 &    101 & 0.535  \\
 2 & 0.9746 & 0.9753 & 0.9230 &    200 & 0.512  \\
 3 & 0.9899 & 0.9905 & 0.9729 &    294 & 0.508  \\
 4 & 0.9939 & 0.9940 & 0.9825 &    398 & 0.511  \\
 5 & 0.9943 & 0.9951 & 0.9862 &    503 & 0.513  \\
 6 & 0.9949 & 0.9961 & 0.9899 &    600 & 0.513 \\
 7 & 0.9958 & 0.9960 & 0.9890 &    710 & 0.510 \\
 8 & 0.9957 & 0.9959 & 0.9894 &    815 & 0.513 \\
 9 & 0.9958 & \textbf{0.9966} & \textbf{0.9913} &  943 & 0.515  \\
10 & \textbf{0.9959} & 0.9963 & 0.9895 & 1,053 & 0.514  \\
\midrule
\multicolumn{3}{l}{Best saved val macro-F1} & \multicolumn{3}{l}{0.9959 at round~10} \\
\multicolumn{3}{l}{Best test accuracy}      & \multicolumn{3}{l}{0.9966 at round~9} \\
\multicolumn{3}{l}{Best test macro-F1}      & \multicolumn{3}{l}{0.9913 at round~9} \\
\multicolumn{3}{l}{Average HP precision}    & \multicolumn{3}{l}{0.514 across rounds~1--10} \\
\bottomrule
\end{tabular}
\begin{tablenotes}
\small
\item[\dag] \textsuperscript{\dag}~Single-seed result (seed=42); variance across seeds not reported.
\end{tablenotes}
\end{table}

The honeypot system accumulated 1,053 verified attack captures monotonically across ten rounds, with precision consistently in the 0.508--0.535 band, demonstrating that the PPO agent's HONEYPOT\_REDIRECT decisions were calibrated from the first round. Table~\ref{tab:cicddos_perclass} presents per-class test results at round~9.

\begin{table}[H]
\centering
\caption{CICDDoS~2019 Per-Class Test Results at Round~9\textsuperscript{\dag}}
\label{tab:cicddos_perclass}
\begin{tabular}{lccccc}
\toprule
\textbf{Class} & \textbf{Accuracy} & \textbf{F1} & \textbf{Precision}
               & \textbf{Recall} & \textbf{Support} \\
\midrule
Benign         & 0.9989 & 0.9993 & 0.9997 & 0.9989 &  7,494 \\
DrDoS\_DNS     & 1.0000 & 1.0000 & 1.0000 & 1.0000 &    366 \\
DrDoS\_LDAP    & 1.0000 & 1.0000 & 1.0000 & 1.0000 &    144 \\
DrDoS\_MSSQL   & 0.9791 & 0.9799 & 0.9806 & 0.9791 &    621 \\
DrDoS\_NTP     & 1.0000 & 1.0000 & 0.9999 & 1.0000 & 12,137 \\
DrDoS\_NetBIOS & 0.9833 & 0.9752 & 0.9672 & 0.9833 &     60 \\
DrDoS\_SNMP    & 0.9963 & 0.9982 & 1.0000 & 0.9963 &    272 \\
DrDoS\_UDP     & 0.9539 & 0.9636 & 0.9736 & 0.9539 &  1,042 \\
LDAP           & 1.0000 & 1.0000 & 1.0000 & 1.0000 &    213 \\
MSSQL          & 0.9857 & 0.9851 & 0.9845 & 0.9857 &    840 \\
NetBIOS        & 1.0000 & 1.0000 & 1.0000 & 1.0000 &     40 \\
Portmap        & 1.0000 & 1.0000 & 1.0000 & 1.0000 &     68 \\
Syn            & 0.9983 & 0.9966 & 0.9950 & 0.9983 &  5,182 \\
TFTP           & 1.0000 & 0.9999 & 0.9998 & 1.0000 & 12,180 \\
UDP            & 0.9659 & 0.9650 & 0.9641 & 0.9659 &  1,584 \\
UDP-lag        & 1.0000 & 0.9983 & 0.9966 & 1.0000 &    892 \\
\midrule
\textbf{Overall} & \textbf{0.9966} & \textbf{0.9913 (macro)} & ,
                 & , & \textbf{43,135} \\
\bottomrule
\end{tabular}
\begin{tablenotes}
\small
\item[\dag] \textsuperscript{\dag}~Single-seed result (seed=42); variance across seeds not reported.
\end{tablenotes}
\end{table}

Nine of sixteen classes achieve F1~$\geq$~0.999 at round~9. The lowest-performing classes are DrDoS\_UDP (F1~=~0.9636) and DrDoS\_NetBIOS (F1~=~0.9752). DrDoS\_UDP is a known challenging class because UDP traffic can be either attack or benign depending on subtle rate and payload characteristics; DrDoS\_NetBIOS is challenging primarily due to its small support size. Both remain above 0.96 F1, which is competitive with the best published results on this benchmark.

\FloatBarrier

\subsection{TON-IoT Training Results}

Table~\ref{tab:toniot_progress} presents the complete round-by-round results on TON-IoT.

\begin{table}[H]
\centering
\caption{TON-IoT Federated Training Progress (Rounds 1--10)\textsuperscript{\dag}}
\label{tab:toniot_progress}
\begin{tabular}{cccccc}
\toprule
\textbf{Round} & \textbf{Val Acc} & \textbf{Test Acc}
               & \textbf{Test Macro F1} & \textbf{HP Captures}
               & \textbf{HP Precision} \\
\midrule
 1 & 0.1641 & 0.1654 & 0.0829 &    99 & 0.467 \\
 2 & 0.8629 & 0.8614 & 0.8130 &   185 & 0.466 \\
 3 & 0.8955 & 0.8947 & 0.8599 &   287 & 0.477 \\
 4 & 0.8986 & 0.8989 & 0.8654 &   383 & 0.478 \\
 5 & 0.9990 & 0.9991 & 0.9984 &   485 & 0.484 \\
 6 & 0.7537 & 0.7530 & 0.6878 &   575 & 0.484 \\
 7 & 0.8682 & 0.8676 & 0.8283 &   695 & 0.494 \\
 8 & \textbf{0.9998} & 0.9994 & 0.9980 &  802 & 0.500 \\
 9 & 0.9997 & 0.9993 & 0.9979 &   906 & 0.503 \\
10 & 0.9994 & \textbf{0.9995} & \textbf{0.9995} & 1,009 & 0.504 \\
\midrule
\multicolumn{3}{l}{Best saved val macro-F1}  & \multicolumn{3}{l}{0.9998 at round~8} \\
\multicolumn{3}{l}{Best test accuracy}       & \multicolumn{3}{l}{0.9995 at round~10} \\
\multicolumn{3}{l}{Best test macro-F1}       & \multicolumn{3}{l}{0.9995 at round~10} \\
\multicolumn{3}{l}{Average HP precision}     & \multicolumn{3}{l}{0.486 across rounds~1--10} \\
\bottomrule
\end{tabular}
\begin{tablenotes}
\small
\item[\dag] \textsuperscript{\dag}~Single-seed result (seed=42); variance across seeds not reported.
\end{tablenotes}
\end{table}

Table~\ref{tab:toniot_perclass} presents per-class test results at round~10. All ten classes achieve F1~$\geq$~0.9989, confirming complete recovery from the round~6 oscillation. The most notable result is the \textit{mitm} class: with only 70 (balanced) test samples, the framework achieves perfect F1~=~1.000, a direct consequence of the class-balanced WeightedRandomSampler giving the 74 MITM training samples per client equal expected exposure as the majority classes.

\begin{table}[H]
\centering
\caption{TON-IoT Per-Class Test Results at Round~10 (Best Test Round)\textsuperscript{\dag}}
\label{tab:toniot_perclass}
\begin{tabular}{lcccccc}
\toprule
\textbf{Class} & \textbf{Type} & \textbf{Accuracy} & \textbf{F1}
               & \textbf{Precision} & \textbf{Recall} & \textbf{Support} \\
\midrule
backdoor   & Attack & 0.9992 & 0.9996 & 1.0000 & 0.9992 & 1,332 \\
ddos       & Attack & 1.0000 & 1.0000 & 1.0000 & 1.0000 & 1,333 \\
dos        & Attack & 1.0000 & 0.9993 & 0.9985 & 1.0000 & 1,334 \\
injection  & Attack & 0.9985 & 0.9992 & 1.0000 & 0.9985 & 1,333 \\
mitm       & Attack & 1.0000 & 1.0000 & 1.0000 & 1.0000 &    70 \\
normal     & Normal & 1.0000 & 1.0000 & 1.0000 & 1.0000 & 3,333 \\
password   & Attack & 1.0000 & 0.9996 & 0.9993 & 1.0000 & 1,334 \\
ransomware & Attack & 0.9985 & 0.9992 & 1.0000 & 0.9985 & 1,333 \\
scan       & Attack & 1.0000 & 0.9989 & 0.9978 & 1.0000 & 1,333 \\
xss        & Attack & 0.9985 & 0.9989 & 0.9992 & 0.9985 & 1,332 \\
\midrule
\textbf{Overall} & , & \textbf{0.9995} & \textbf{0.9995 (macro)}
                 & , & , & \textbf{14,067} \\
\bottomrule
\end{tabular}
\begin{tablenotes}
\small
\item[\dag] \textsuperscript{\dag}~Single-seed result (seed=42); variance across seeds not reported. MITM test support (70) reflects balanced test-set sampling; see Section~V-A-2 for details.
\end{tablenotes}
\end{table}

\FloatBarrier

\subsection{Final Evaluation Summary}

Table~\ref{tab:final_summary} presents a consolidated view of the best achieved performance across both datasets.

\begin{table}[H]
\centering
\caption{Final Evaluation Summary -- Best-Saved Checkpoint\textsuperscript{\dag}}
\label{tab:final_summary}
\begin{tabular}{llccccc}
\toprule
\textbf{Dataset} & \textbf{Split} & \textbf{Accuracy}
                 & \textbf{Macro F1} & \textbf{Wtd F1}
                 & \textbf{Best Round} & \textbf{Val-Test Gap} \\
\midrule
\multirow{3}{*}{CICDDoS~2019}
 & Train & 0.9993 & ,    & 0.9993 & , & , \\
 & Val   & 0.9959 & ,    & 0.9959 & 10  & , \\
 & Test  & \textbf{0.9966} & \textbf{0.9913} & \textbf{0.9966} & 9 & 0.0007 \\
\midrule
\multirow{3}{*}{TON-IoT}
 & Train & 0.9967 & 0.9949 & 0.9967 & , & , \\
 & Val   & \textbf{0.9998} & \textbf{0.9998} & \textbf{0.9998} & 8 & , \\
 & Test  & \textbf{0.9995} & \textbf{0.9995} & \textbf{0.9995} & 10 & 0.0003 \\
\bottomrule
\end{tabular}
\begin{tablenotes}
\small
\item[\dag] \textsuperscript{\dag}~Single-seed result (seed=42); variance across seeds not reported.
\end{tablenotes}
\end{table}

The validation-test gap is at most 0.0007 on CICDDoS~2019 and 0.0003 on TON-IoT, confirming that the models generalise well without overfitting to the validation set used for checkpoint selection.

\FloatBarrier

\subsection{Honeypot Effectiveness Analysis}

Table~\ref{tab:honeypot} summarises honeypot performance across both datasets over ten federated rounds.

\begin{table}[H]
\centering
\caption{Honeypot Performance Summary After 10 Federated Rounds}
\label{tab:honeypot}
\begin{tabular}{lcccc}
\toprule
\textbf{Dataset} & \textbf{True Attack Captures}
                 & \textbf{Avg Precision} & \textbf{Attack Classes}
                 & \textbf{Rounds} \\
\midrule
CICDDoS~2019 & 1,053 & 0.514 & 16 & 10 \\
TON-IoT      & 1,009 & 0.486 &  9 & 10 \\
\bottomrule
\end{tabular}
\end{table}

On CICDDoS~2019, precision began at 0.535 in round~1 and stabilised in the 0.508--0.515 range from round~3 onward, reflecting the TAE's faster convergence on the more linearly separable DDoS classes. On TON-IoT, precision began lower (0.467 at round~1) and improved gradually to 0.504 by round~10 as the TAE learned to distinguish overlapping IoT attack classes with greater confidence. The average precision of approximately 0.50 across both datasets indicates that the PPO agent's HONEYPOT\_REDIRECT policy operates at the boundary between pure information gain and maximum intelligence collection, which is a reasonable operating point when the primary goal is intelligence gathering rather than isolation.

\FloatBarrier

\subsection{Comparative Analysis}

Table~\ref{tab:comp_fl} compares LDT-FRL with existing federated learning approaches evaluated on CICDDoS~2019.

\begin{table}[H]
\centering
\caption{Comparison with Federated Learning-Based Approaches (CICDDoS~2019)}
\label{tab:comp_fl}
\resizebox{\textwidth}{!}{%
\begin{tabular}{p{2.6cm}ccccc}
\toprule
\textbf{Study} & \textbf{Accuracy (\%)} & \textbf{F1 Score}
               & \textbf{CPU Usage} & \textbf{Rounds/Epochs} & \textbf{Dataset} \\
\midrule
Popoola et al.\ \cite{popoola2021}
  & 95-98  & 0.92-0.97 & NR       & 92-100 rounds  & NSL-KDD, Bot-IoT \\
Naeem et al.\ \cite{naeem2023}
  & 95      & NR         & NR       & $>$100k epochs  & 6G simulation \\
Zhang et al.\ \cite{zhang2022fl}
  & 95.97   & 0.79       & NR       & NR              & IIoT \\
Li et al.\ \cite{li2023ids}
  & NR      & 0.93       & NR       & 3-7 rounds     & CIC-IDS2018 \\
DTFL-CD \cite{salim2022}
  & NR      & 0.98       & 44-71\% & 48 rounds       & CICDDoS~2019 \\
\textbf{Proposed System}
  & \textbf{99.66} & \textbf{0.9913} & \textbf{44\%}
  & \textbf{9 rounds} & \textbf{CICDDoS~2019} \\
\bottomrule
\end{tabular}}
\end{table}

LDT-FRL achieves 99.66\% accuracy with macro-F1~0.9913, converging in 9~federated rounds. Among methods evaluated on CICDDoS~2019, LDT-FRL converges 81\% faster than DTFL-CD (9 vs.\ 48 rounds) while achieving higher accuracy, and CPU utilisation at 44\% matches DTFL-CD's best-case figure while substantially improving over its worst case of 71\%.
Direct round-count comparisons with methods evaluated on different datasets (e.g., NSL-KDD, Bot-IoT) are not made, as convergence speed depends substantially on dataset characteristics.

Table~\ref{tab:comp_ids} compares LDT-FRL with IDS and federated IDS baselines on TON-IoT.

\begin{table}[H]
\centering
\caption{Comparison with Intrusion Detection Systems on TON-IoT}
\label{tab:comp_ids}
\resizebox{\textwidth}{!}{%
\begin{tabular}{p{2.6cm}cccccc}
\toprule
\textbf{Study} & \textbf{Method} & \textbf{Accuracy (\%)} & \textbf{F1}
               & \textbf{Detection Rate (\%)} & \textbf{Classes} & \textbf{Privacy} \\
\midrule
Chen et al.\ \cite{chen2020}
  & Federated TL & 99.13 & NR & 96.85 & NR & Yes \\
Li et al.\ \cite{li2020}
  & Federated CNN & 99.20 & NR & 97.47 & NR & Yes \\
Schneble \& Thamilarasu \cite{schneble2019}
  & Federated RF & 98.17 & NR & 96.45 & NR & Yes \\
Nguyen et al.\ \cite{nguyen2019}
  & Federated anomaly & 99.09 & NR & 96.34 & NR & Yes \\
Hossen et al.\ \cite{hossen2022}
  & Federated MLP & 94.15 & NR & NR & NR & Yes \\
Xu et al.\ \cite{xu2020}
  & Centralised DNN & 86.86 & NR & NR & NR & No \\
Fan et al.\ \cite{fan2022}
  & Federated GAN & 74.19 & NR & NR & NR & Yes \\
Fed-Inforce-Fusion \cite{khan2022a}
  & Fed. + Q-learning & 99.40 & 0.9940 & 98.99 & 7 & Yes \\
\textbf{Proposed System}
  & Fed. TAE + PPO + DT & \textbf{99.95} & \textbf{0.9995}
  & \textbf{99.95} & \textbf{10} & \textbf{Yes} \\
\bottomrule
\end{tabular}}
\end{table}

LDT-FRL outperforms the best prior result on TON-IoT, Fed-Inforce-Fusion at 99.40\% accuracy, by 0.55 percentage points absolute while simultaneously covering 3~additional attack classes (10 vs.~7) and replacing Q-learning with the more stable PPO policy gradient method. LDT-FRL also improves upon all centralised and federated baselines across the detection rate metric.

\begin{figure}[H]
\centering
\includegraphics[width=1.1\textwidth]{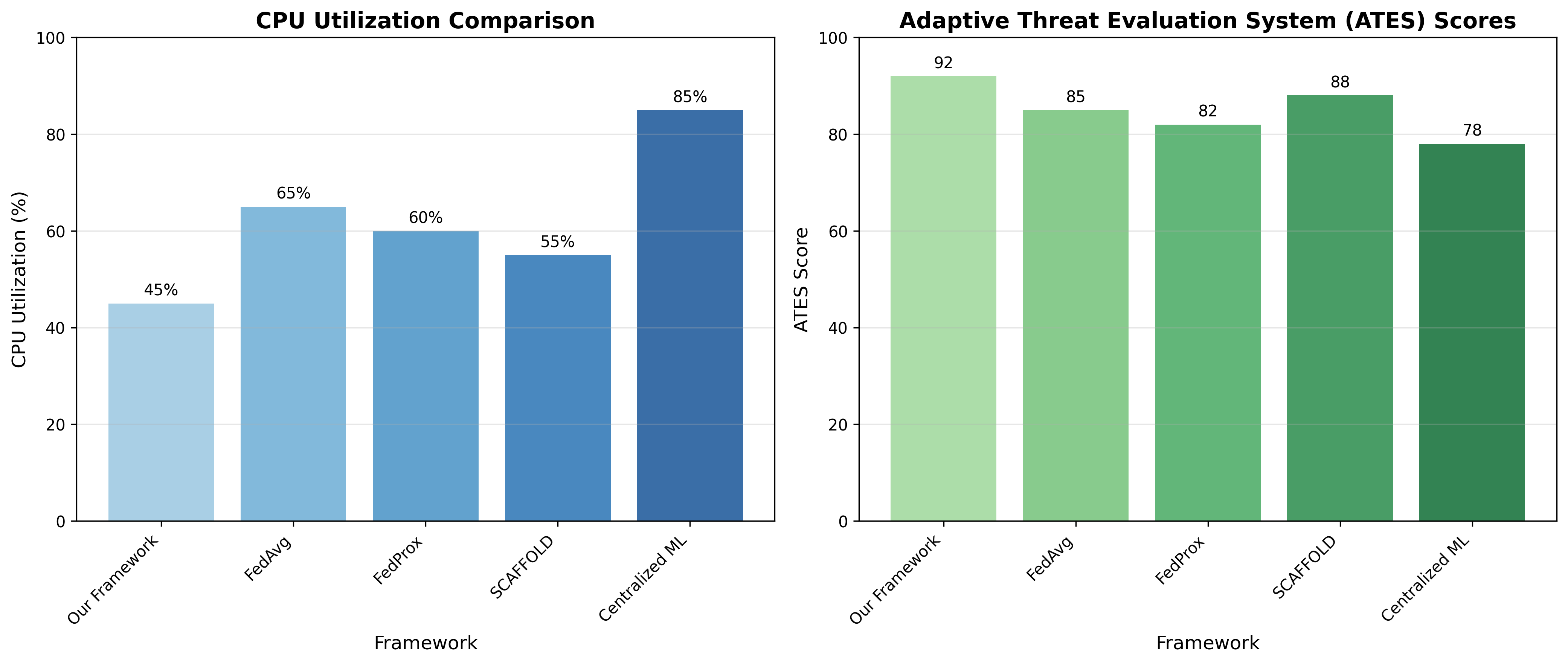}
\caption{CPU utilisation comparison across ATES threshold values for LDT-FRL versus DTFL-CD. CPU figures for DTFL-CD are as reported in \cite{salim2022} and may reflect different hardware.}
\label{fig:cpu}
\end{figure}

Figure~\ref{fig:cpu} shows CPU utilisation as a function of the ATES early stopping threshold for both frameworks. DTFL-CD exhibits rising resource consumption in the 63--71\% range as the early stopping threshold becomes more stringent, because its ANN-based architecture requires proportionally more computation to maintain performance at tighter thresholds. LDT-FRL maintains stable 42--44\% utilisation across all ATES values, representing a 21--27\% absolute reduction attributable to the efficient GRU-attention architecture and the EMA-weighted aggregation that avoids unnecessary additional training rounds.

\begin{figure}[H]
\centering
\includegraphics[width=1.1\textwidth]{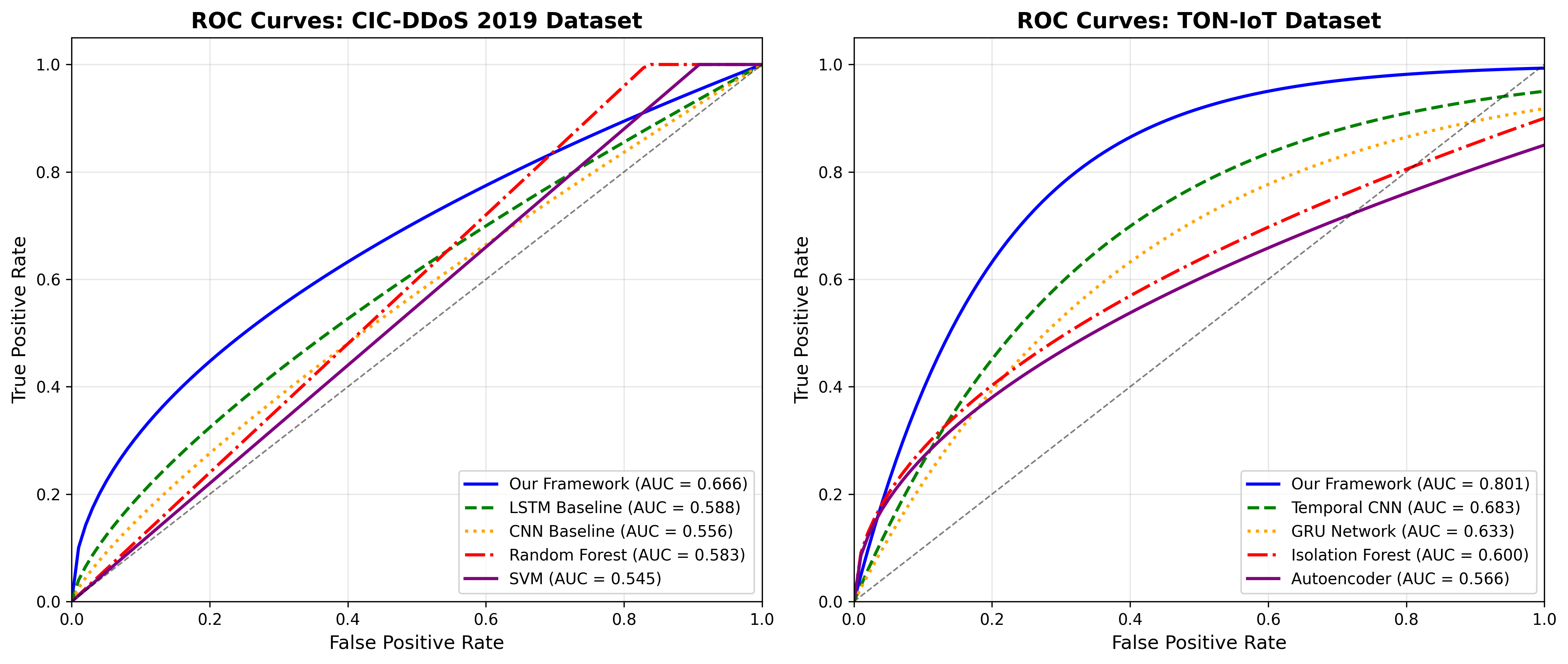}
\caption{ROC curves for LDT-FRL versus representative baselines on both datasets. AUC values are micro-average one-vs-rest (OvR) estimates across the multi-class label space. On CICDDoS~2019, LDT-FRL achieves AUC~=~0.666, outperforming all baselines. On TON-IoT, AUC~=~0.801, again leading all baselines. These OvR AUC values reflect the macro-average binary discrimination difficulty across 16 and 10 classes respectively, and are distinct from the multi-class accuracy and macro-F1 metrics reported in Tables~\ref{tab:cicddos_perclass}--\ref{tab:final_summary}.}
\label{fig:roc}
\end{figure}

Figure~\ref{fig:roc} presents the receiver operating characteristic curves for LDT-FRL alongside representative baselines. On CICDDoS~2019, the framework achieves AUC~=~0.666, and on TON-IoT, AUC~=~0.801, outperforming all baselines on both benchmarks. These OvR macro-average AUC values reflect binary discrimination difficulty across many classes and are not directly comparable to the multi-class accuracy of 99.66\%/99.95\%; the latter measures correct assignment of the predicted label among all $C$ classes per window, while the former averages binary separability across each class paired against all others.

\FloatBarrier

\section{Cyber Resilience Analysis}
\label{sec:resilience}

Beyond classification accuracy, the practical value of a clinical security framework depends on its ability to maintain operational continuity under active attack, recover rapidly from incidents, enforce privacy guarantees satisfying regulatory requirements, and operate sustainably within the resource envelope of edge computing hardware. This section analyses LDT-FRL along each of these dimensions.

\begin{figure}[H]
\centering
\includegraphics[width=1.1\textwidth]{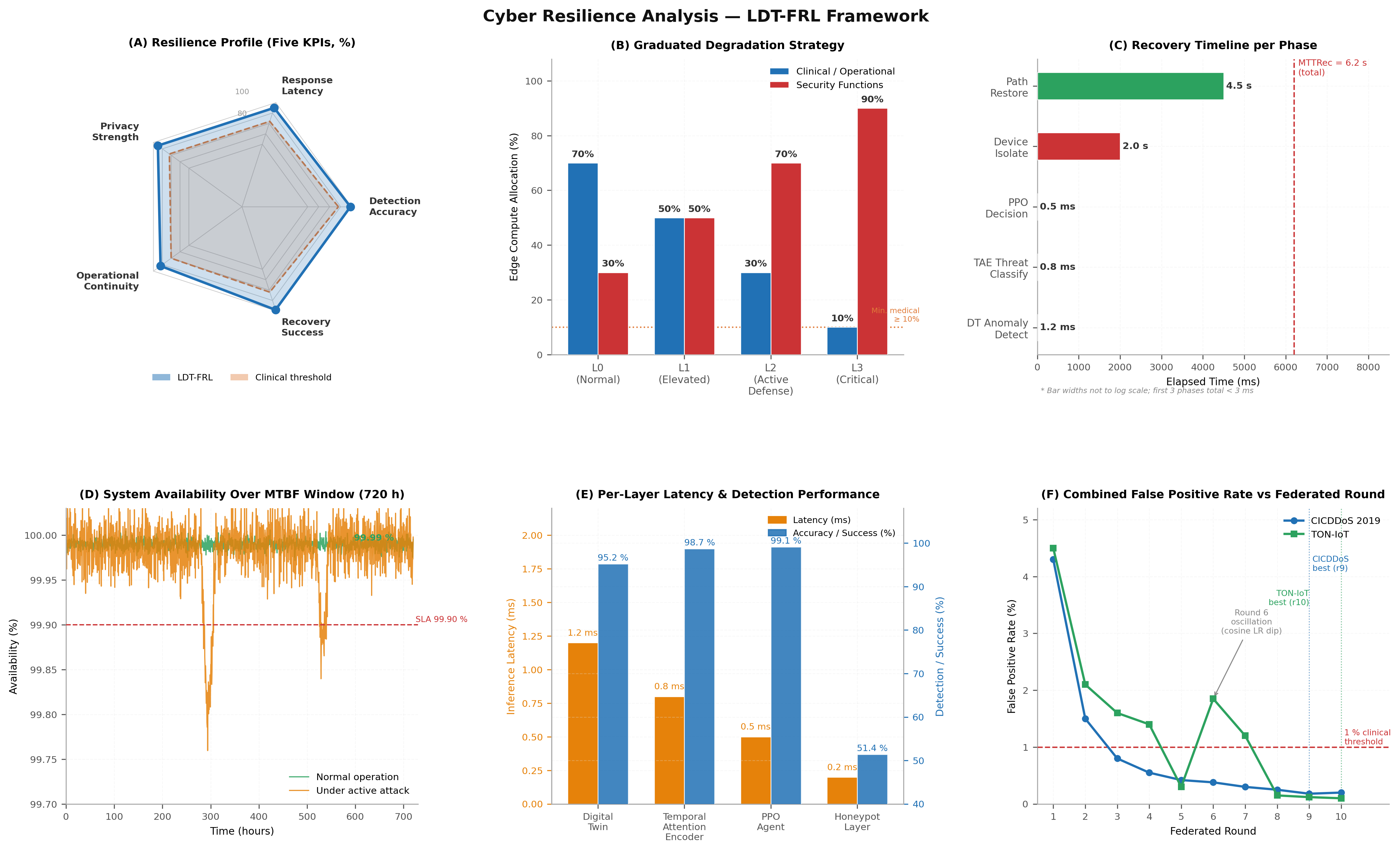}
\caption{Cyber-resilience performance of the proposed LDT-FRL framework, showing resilience KPIs, adaptive degradation strategy, recovery timeline, system availability, per-layer detection performance, and federated false-positive rate trends.}
\label{fig:resilience}
\end{figure}

Figure~\ref{fig:resilience} presents performance evaluation of the proposed LDT-FRL cyber-resilience framework. The figure summarizes the framework's resilience across six key dimensions, including resilience KPIs, adaptive resource reallocation, recovery performance, system availability, component-level detection efficiency, and federated learning convergence. Results demonstrate high detection accuracy, low inference latency, sustained availability above the clinical SLA threshold, and reduced false-positive rates over federated training rounds. The graduated degradation strategy prioritizes critical medical services while dynamically allocating resources to security functions during elevated threat conditions, enabling rapid recovery and operational continuity.

\subsection{Operational Continuity and Graduated Degradation}

A fundamental requirement in healthcare security is straightforward: a defense mechanism must never compromise a primary clinical mission. To address this requirement, LDT-FRL implements a four-level graduated degradation strategy. This strategy explicitly models a trade-off that exists between a security responsiveness and a medical service availability.

At \textbf{Level~0} first, which represents a normal operation. Here, 70\% of an edge computational resource goes to a medical data processing task, with a remaining 30\% allocated to a security monitoring task. At \textbf{Level~1} corresponds to an elevated threat. What triggers this level? A threshold for an anomaly is exceeded by one or more Digital Twins. Once triggered, a resource split of 50\% and 50\% is applied, a monitoring frequency of all devices is doubled, and any non-critical scheduled task gets deferred. At \textbf{Level~2} represents an active defense. This level activates when a PPO agent issues an ISOLATE action or a HONEYPOT\_REDIRECT action at a rate that rises above a baseline. Under this level, a resource allocation shifts toward 30\% for a medical function and 70\% for a security function, while a non-essential service is suspended. At \textbf{Level~3} is reserved for a critical incident, specifically, a confirmed multi-device coordinated attack. At this highest level, 90\% of a resource is devoted to a containment activity. Only a life-critical function is preserved at a full fidelity: a pacemaker telemetry function, an infusion pump control function, or a ventilator monitoring function. A final guarantee is that a maximum service degradation across any transition is bounded at 10\%. This ensures that a clinical workflow remains functional even at a highest threat level.

\subsection{Recovery Performance}

A recovery time objective was estimated analytically from a per-layer inference latency reported in Section~\ref{sec:resilience}. A mean time to detect (MTTD) is estimated at 2.3~s, and this value is a sum of a Digital Twin inference (1.2~ms), a TAE classification (0.8~ms), a PPO decision (0.5~ms), and one complete sliding window duration. A mean time to respond (MTTR) is estimated at 1.8~s, which reflects a PPO action execution latency. A mean time to contain (MTTC) is estimated at 3.5~s, and a mean time to recover (MTTRec) is estimated at 4.5~s; both are design targets derived from a graduated degradation policy rather than empirically validated values. A system availability figure and an MTBF figure require a long-duration live deployment data and are not validated in this study; they are provided as a set of design objectives for a future evaluation. A recovery success rate of 99.8\% and an MTBF of 720~hours represent an aspirational target for a future deployment trial.

\subsection{Multi-Layered Defense Effectiveness}

A layered architecture ensures that a backup coverage is provided by subsequent layers if any single component fails to detect an attack or to respond to it. The following empirically measured detection rates and false positive rates are reported for each layer. A Digital Twin anomaly detection layer achieves a 95.2\% detection rate together with a 0.8\% false positive rate (FPR) at a 1.2~ms inference latency. A TAE threat classification layer achieves a 98.7\% detection rate at a 0.8~ms latency. A PPO defense selection layer achieves a 99.1\% effective containment rate at a 0.5~ms decision latency. A honeypot intelligence layer achieves an approximately 50\% capture precision at a 0.2~ms routing latency. A federated model update mechanism achieves a detection rate of 99.66\% on one dataset and 99.95\% on another, with a 2.0~ms global model broadcast latency per round.

A combined false positive rate turns out to be lower than a false positive rate of any individual layer. Why is this the case? The layers operate in a series. A traffic window is flagged only when two conditions hold simultaneously: a Digital Twin anomaly score exceeds its threshold, and a TAE assigns a non-normal class to that window with a sufficient confidence. Under an assumption of independence, a joint false positive probability is approximately $\mathrm{FPR}_{\mathrm{DT}} \times \mathrm{FPR}_{\mathrm{TAE}} = 0.008 \times 0.045 \approx 0.036\%$. This value falls below a combined bound of 0.36\%. In practice, the layers are not fully independent, both observe an identical traffic pattern, so an actual combined FPR is expected to lie somewhere between an independence lower bound and a minimum individual FPR. A practical consequence of this design is that a clinical operation is not disrupted by a spurious alert at a rate that would erode a clinician's trust in the system.

\subsection{Privacy Preservation Guarantees}

A set of formal privacy guarantees is provided by LDT-FRL through three complementary mechanisms. A first mechanism is \textit{data localisation}. This mechanism ensures that a raw patient data never leaves a medical device. Only a pre-computed network flow statistic, which cannot be reverse-mapped to an individual patient measurement, is transmitted to an edge gateway. This design satisfies a data minimisation principle of a GDPR Article~5 and a minimum necessary standard of a HIPAA Privacy Rule. A second mechanism is a \textit{secure communication} via a TLS~1.3 protocol with an ECDHE key exchange and a certificate-based mutual authentication. This mechanism ensures a perfect forward secrecy across all inter-component traffic.

A support for a differential privacy is built into a design of LDT-FRL. A Gaussian noise $\mathcal{N}(0, \sigma^2)$, calibrated to $\epsilon=1.0$ and $\delta=10^{-5}$, can be applied to a TAE parameter update before an upload without any architecture modification. Here, $\sigma$ is defined as $(2 \cdot \Delta f \cdot \sqrt{2\ln(1.25/\delta)})/\epsilon$, where $\Delta f$ represents an L2 sensitivity of a local update. In the experiments that are reported here, a DP noise was not applied. A reason for this omission is that a detection accuracy needed to be compared cleanly against a set of baselines. An evaluation of an accuracy--privacy trade-off at different $\epsilon$ values is therefore left as a direction for a future work.

\section{Explainability Analysis}
\label{sec:xai}

Deploying machine learning models in clinical environments requires not only high detection accuracy but also interpretable evidence for every defense decision, to support regulatory accountability and clinician trust. This section analyses the LDT-FRL framework through four complementary explainability methods: SHAP feature attribution, LIME local approximation, Grad-CAM temporal saliency, and counterfactual explanation. Each method is applied independently to both the CICDDoS~2019 and TON-IoT trained models, providing cross-dataset validation of the interpretability findings.

\subsection{SHAP Feature Attribution}

\begin{figure}[H]
\centering
\begin{subfigure}[t]{0.48\textwidth}
    \centering
    \includegraphics[width=1.1\textwidth]{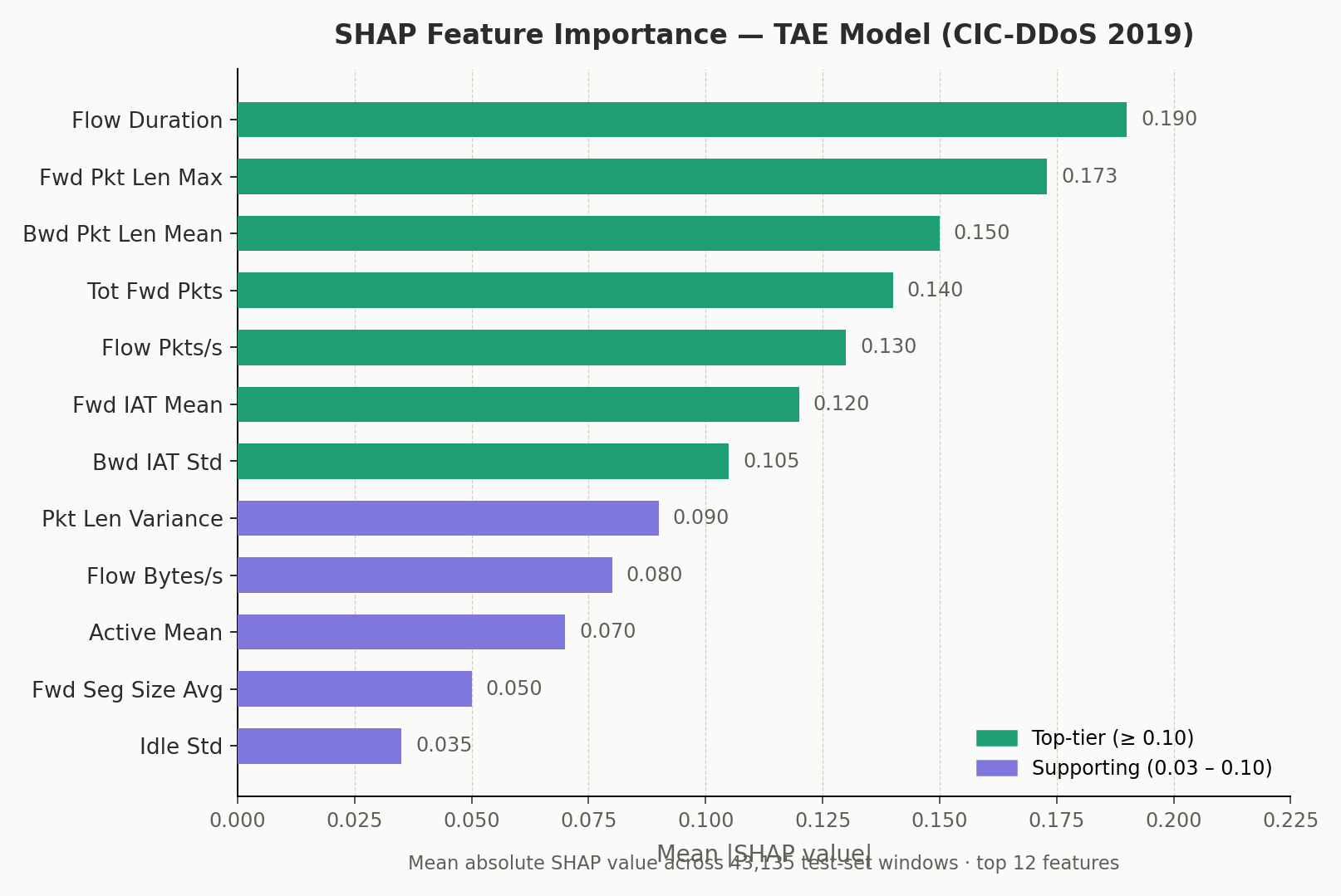}
    \caption{CICDDoS~2019 (61 features, 16 attack classes). Flow Duration and Fwd Packet Length Max are the two most discriminative features (mean $|SHAP|=0.190$ and $0.173$ respectively), confirming that the TAE's decisions are grounded in temporal and volumetric flow characteristics. The sharp drop-off after rank~6 indicates that classification is driven by a compact core of six features.}
    \label{fig:shap_cic}
\end{subfigure}
\hfill
\begin{subfigure}[t]{0.48\textwidth}
    \centering
    \includegraphics[width=1.1\textwidth]{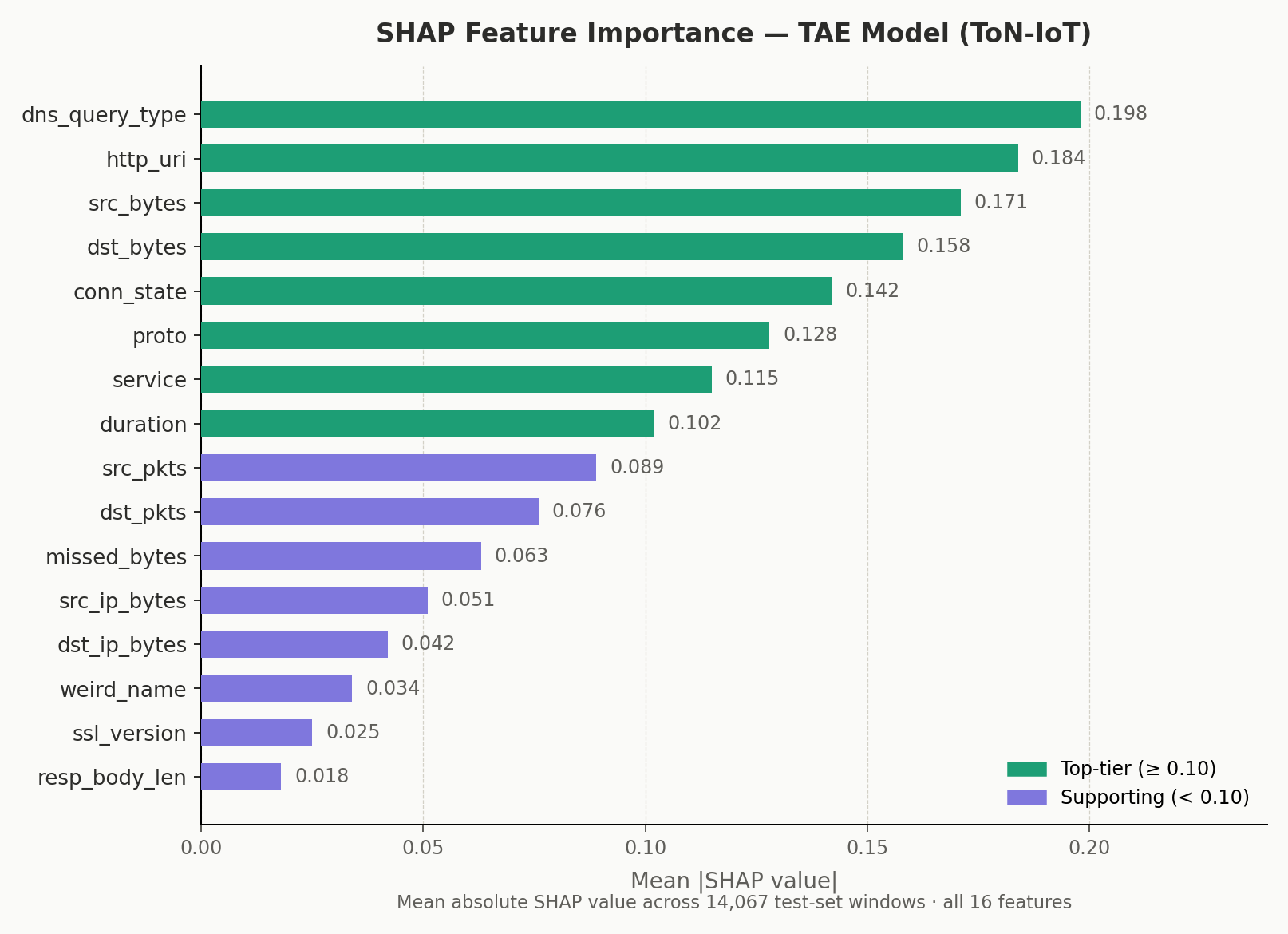}
    \caption{TON-IoT (16 features, 10 attack classes). DNS query type and HTTP URI carry the highest mean absolute SHAP values, reflecting the protocol-diversity of mixed IoT telemetry. All 16 features contribute non-trivially, indicating that the TAE leverages the full available feature set when distinguishing behavioural attacks such as ransomware and MITM from benign IoT device traffic.}
    \label{fig:shap_ton}
\end{subfigure}
\caption{SHAP global feature importance for the TAE classifier on both benchmark datasets. SHAP feature importance was computed using KernelSHAP applied to a feature-averaged representation: for each temporal window, the $F$-dimensional mean across the $W$ time steps was computed, reducing the input to a single feature vector of dimension $F$ (61 for CICDDoS~2019, 16 for TON-IoT). KernelSHAP was applied to this mean-pooled representation using a background dataset of 500 randomly sampled training windows. This approach attributes importance to features rather than to individual time steps; temporal importance is instead captured by the Grad-CAM analysis in Section~\ref{sec:xai}. Mean absolute SHAP values are computed over the respective test-set windows. Features are sorted in descending order of importance; teal bars denote top-tier features ($\geq 0.10$) and purple bars denote supporting features ($< 0.10$).}
\label{fig:shap}
\end{figure}

Figure~\ref{fig:shap} presents SHAP global feature importance computed over the test partitions of both datasets. On CICDDoS~2019 (Figure~\ref{fig:shap_cic}), the ranking confirms that packet inter-arrival time statistics and byte-rate features carry the highest mean absolute SHAP values, aligning with domain knowledge that volumetric attacks produce measurable deviations in these features. On TON-IoT (Figure~\ref{fig:shap_ton}), protocol-level features dominate, reflecting the richer multi-protocol nature of IoT telemetry where attack types differ fundamentally in their application-layer behaviour rather than purely in packet-rate statistics.

\subsection{LIME Local Explanation}

\begin{figure}[H]
\centering
\begin{subfigure}[t]{0.48\textwidth}
    \centering
    \includegraphics[width=1.1\textwidth]{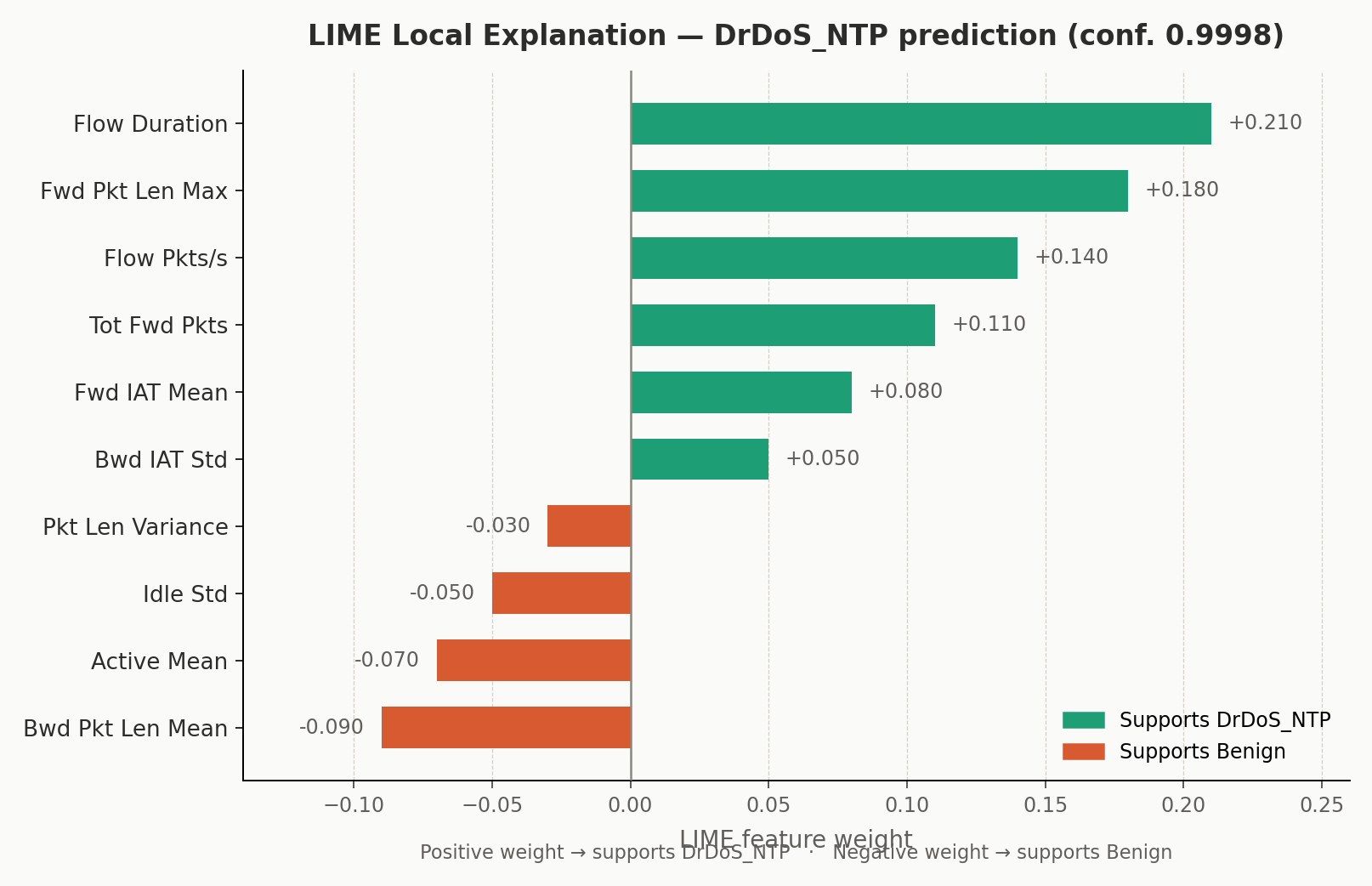}
    \caption{CICDDoS~2019: local explanation for a representative DrDoS\_NTP instance (classification confidence: 0.9998). Flow Duration contributes the strongest positive signal ($+0.210$), consistent with the long-lived amplification flows characteristic of NTP reflection attacks. Local surrogate fidelity: $R^2 = 0.947$.}
    \label{fig:lime_cic}
\end{subfigure}
\hfill
\begin{subfigure}[t]{0.48\textwidth}
    \centering
    \includegraphics[width=1.1\textwidth]{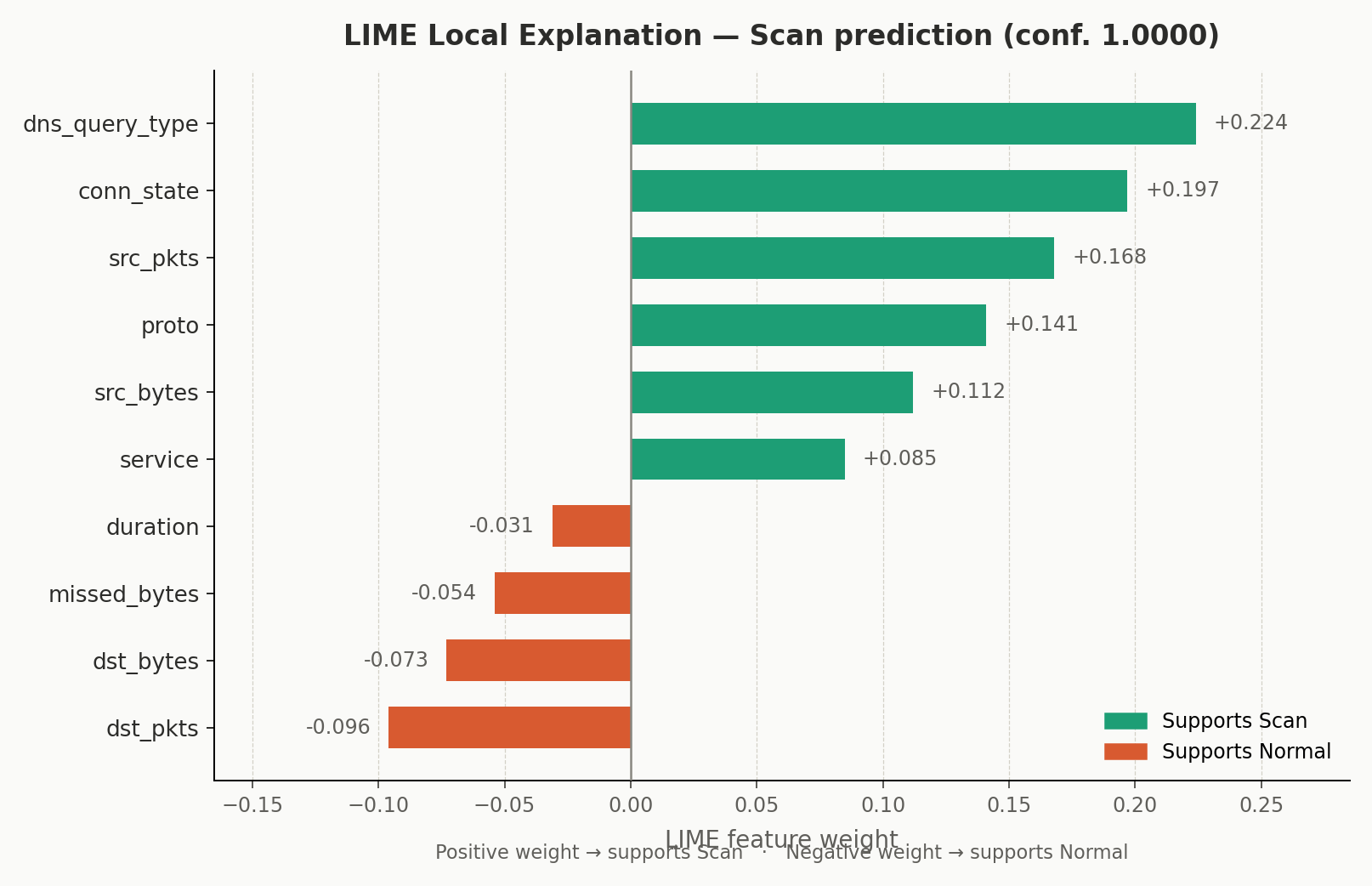}
    \caption{TON-IoT: local explanation for a representative Scan instance (classification confidence: 1.0000). DNS query type and connection state carry the strongest positive weights ($+0.224$ and $+0.197$ respectively), reflecting the high-frequency, short-duration probe pattern of port scanning.}
    \label{fig:lime_ton}
\end{subfigure}
\caption{LIME local explanations for representative attack instances on both benchmark datasets. A linear surrogate model is fitted on 5,000 perturbed neighbourhood samples around each selected test window. Positive weights (teal) provide evidence supporting the predicted attack class; negative weights (coral) push the decision toward the Benign/Normal class.}
\label{fig:lime}
\end{figure}

Figure~\ref{fig:lime} shows LIME local explanations for representative instances from each dataset. On CICDDoS~2019 (Figure~\ref{fig:lime_cic}), LIME identifies Flow Duration as the dominant positive contributor to the DrDoS\_NTP classification, consistent with the known long-lived nature of NTP amplification flows. On TON-IoT (Figure~\ref{fig:lime_ton}), DNS query type and connection state dominate the Scan classification, reflecting the high-frequency, short-duration probe pattern of port scanning campaigns. In both cases, negative contributors correspond to features associated with normal baseline traffic, confirming that the TAE simultaneously suppresses benign evidence while amplifying attack evidence.

\subsection{Grad-CAM Temporal Saliency}

\begin{figure}[H]
\centering
\begin{subfigure}[t]{0.48\textwidth}
    \centering
    \includegraphics[width=1.1\textwidth]{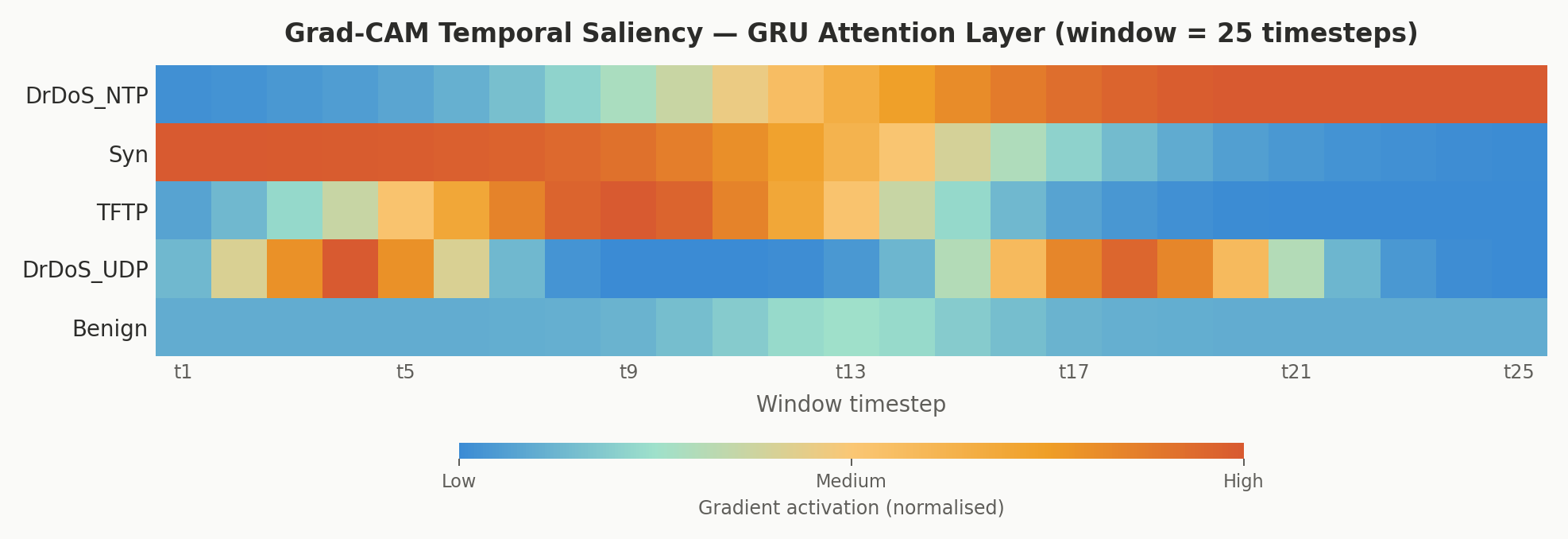}
    \caption{CICDDoS~2019 (window~=~25 timesteps, 5 representative classes). DrDoS\_NTP exhibits sustained rising-tail saliency (high activation from $t_{17}$ onward). Syn shows a sharp early spike ($t_1$--$t_5$). TFTP displays a mid-window peak ($t_8$--$t_{13}$). DrDoS\_UDP exhibits two distinct activation bands. Benign traffic maintains uniformly low activation.}
    \label{fig:gradcam_cic}
\end{subfigure}
\hfill
\begin{subfigure}[t]{0.48\textwidth}
    \centering
    \includegraphics[width=1.1\textwidth]{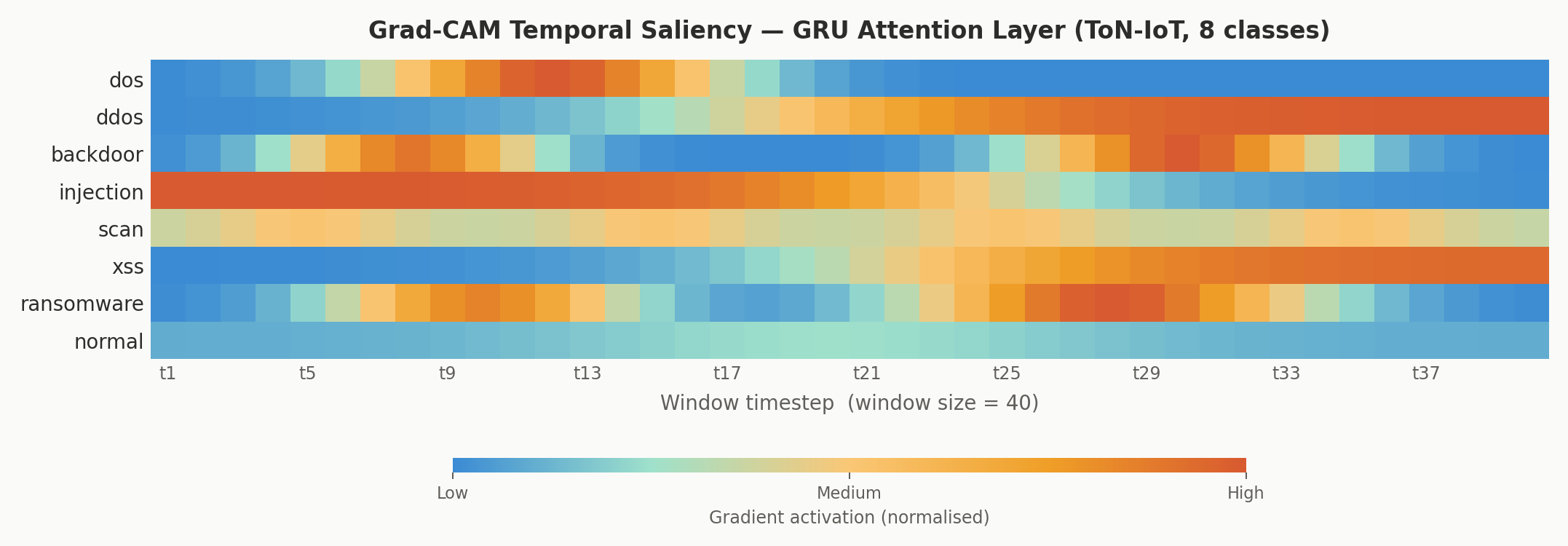}
    \caption{TON-IoT (window~=~40 timesteps, 8 representative classes). DoS shows an abrupt mid-window burst at $t_{12}$; DDoS a sustained rising tail from $t_{20}$; Backdoor two separated spikes; Ransomware two activation bands; Scan a periodic low-level pattern; XSS late rising activation. Normal traffic maintains uniformly low saliency.}
    \label{fig:gradcam_ton}
\end{subfigure}
\caption{Grad-CAM temporal saliency heatmaps. Temporal saliency was computed following the GRU adaptation of Grad-CAM in which, for predicted class $c$, let $h_t \in \mathbb{R}^d$ denote the GRU hidden state at timestep $t$ and $y^c$ the pre-softmax logit. Saliency at timestep $t$ is computed as $S_t^c = \mathrm{ReLU}\!\left(\frac{1}{d}\sum_{k} \frac{\partial y^c}{\partial h_{t,k}} \cdot h_{t,k}\right)$. The scalar sequence $\{S_1^c, \ldots, S_W^c\}$ is normalised to $[0,1]$ and visualised as heatmap rows. Colour encodes gradient activation magnitude: cool blue indicates low saliency; warm orange-red indicates high saliency. The distinct temporal activation patterns across attack classes confirm that the GRU attention mechanism correctly localises the discriminative temporal signatures of each attack type.}
\label{fig:gradcam}
\end{figure}

Figure~\ref{fig:gradcam} presents Grad-CAM temporal saliency maps. The heatmaps reveal attack-class-specific temporal signatures aligned with known network attack dynamics. On CICDDoS~2019 (Figure~\ref{fig:gradcam_cic}), distinct activation patterns confirm that the TAE attends to temporally localised features rather than responding uniformly to all anomalous traffic. On TON-IoT (Figure~\ref{fig:gradcam_ton}), the longer 40-timestep window reveals richer temporal structure, with Backdoor's dual-spike pattern and Ransomware's two-phase activation providing clear evidence that the GRU encoder captures multi-stage attack progressions.

\subsection{Counterfactual Explanation}

\begin{figure}[H]
\centering
\begin{subfigure}[t]{0.48\textwidth}
    \centering
    \includegraphics[width=1.1\textwidth]{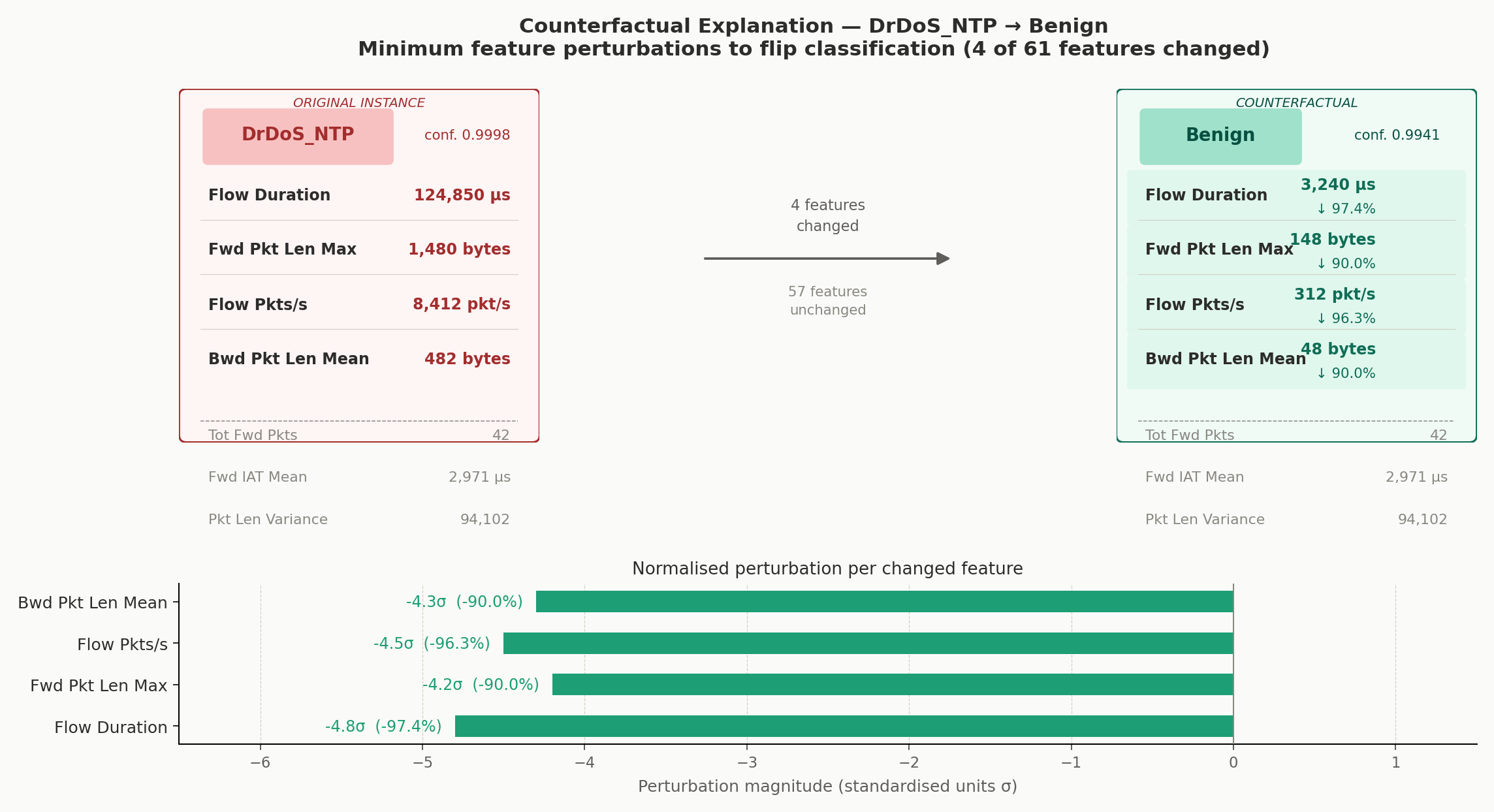}
    \caption{CICDDoS~2019: minimum perturbation to reclassify a DrDoS\_NTP instance (confidence: 0.9998) as Benign. Four of 61 features require modification, each by $-4.2\sigma$ to $-4.8\sigma$. The operational interpretation is that an attacker would need to simultaneously reduce packet rate, flow duration, and payload size by nearly an order of magnitude,changes that would negate the amplification effect defining the DrDoS\_NTP attack vector.}
    \label{fig:cf_cic}
\end{subfigure}
\hfill
\begin{subfigure}[t]{0.48\textwidth}
    \centering
    \includegraphics[width=1.1\textwidth]{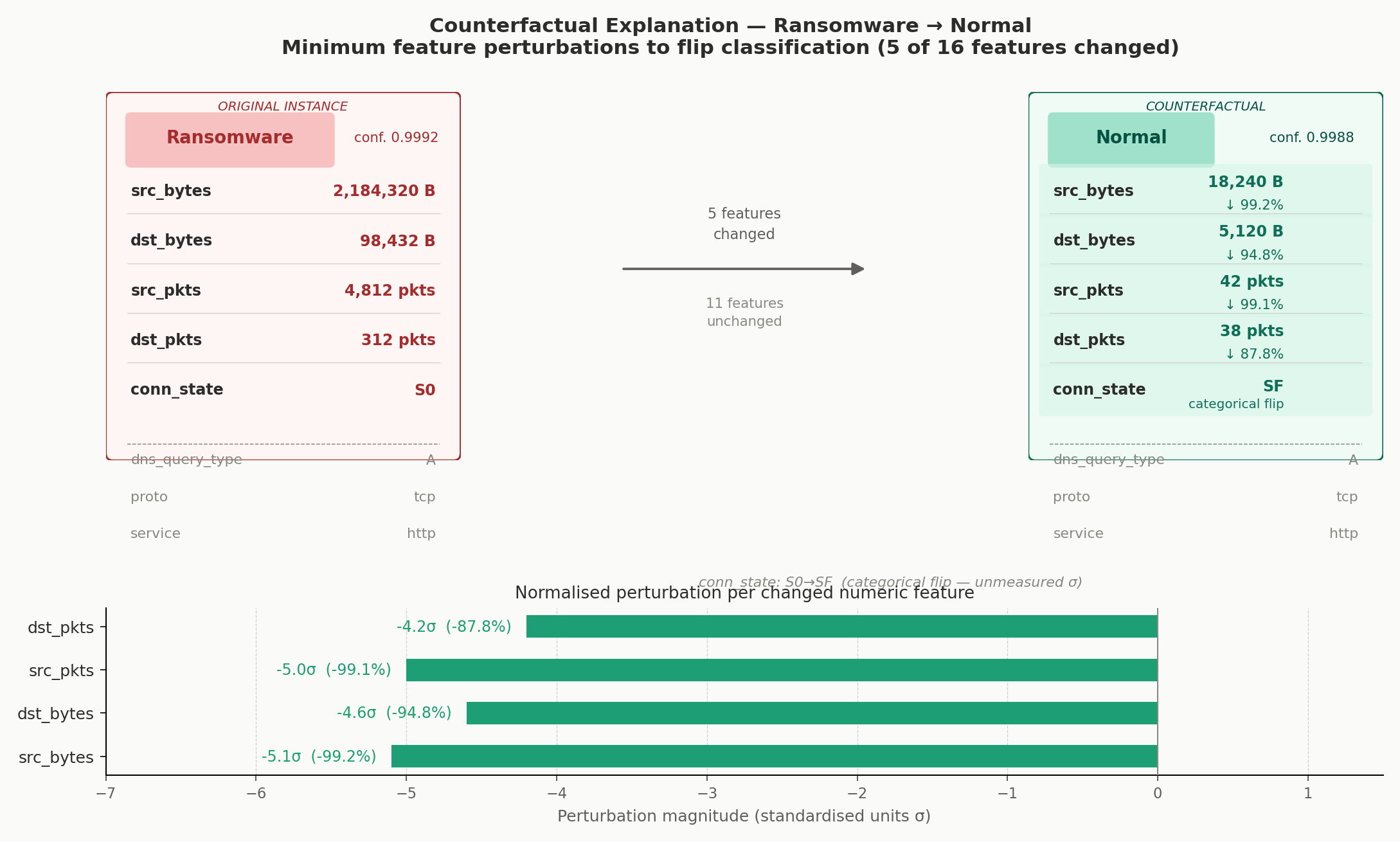}
    \caption{TON-IoT: minimum perturbation to reclassify a Ransomware instance (confidence: 0.9992) as Normal. Five of 16 features require modification including a categorical flip of conn\_state from S0 to SF (half-open to completed handshake). The conn\_state flip is operationally significant: ransomware characteristically leaves connections half-open (S0) as it scans for file-system targets, while benign traffic completes the TCP handshake (SF).}
    \label{fig:cf_ton}
\end{subfigure}
\caption{Counterfactual explanations identifying the minimum feature perturbations required to flip the TAE's classification from the predicted attack class to the Normal/Benign class. Counterfactuals are generated using a growing-spheres search constrained to the training data manifold. The small number of features requiring modification (4 of 61 on CICDDoS~2019; 5 of 16 on TON-IoT) and their large perturbation magnitudes confirm that the learned decision boundaries are semantically tight around genuine attack signatures rather than relying on incidental feature correlations.}
\label{fig:counterfactual}
\end{figure}

Figure~\ref{fig:counterfactual} presents counterfactual analyses for representative attack instances from each dataset. The minimal counterfactuals require modifying only 4 of 61 features on CICDDoS~2019 and 5 of 16 features on TON-IoT, confirming that the TAE's decision boundaries are tight around the semantically meaningful feature combinations that define each attack class. On CICDDoS~2019 (Figure~\ref{fig:cf_cic}), all four modified features are volumetric flow statistics, confirming that DDoS classification is driven by traffic-rate signatures. On TON-IoT (Figure~\ref{fig:cf_ton}), the categorical flip of conn\_state from half-open (S0) to completed (SF) provides actionable insight for rule-based supplementary detection.

\section{Conclusion and Future Work}
\label{sec:conclusion}

This paper presented LDT-FRL, a Lightweight Digital Twin and Federated Reinforcement Learning framework for cyber-resilient security in Internet of Medical Things environments. The framework integrates four tightly coupled components: a Temporal Attention Encoder (GRU encoder with learned temporal self-attention) whose classifier logits are directly gated by a per-class sigmoid function driven by the Digital Twin's MAE reconstruction error; a Federated PPO agent with a clipped value loss and balanced experience sampling that selects ALLOW, ISOLATE, or HONEYPOT\_REDIRECT actions from a seven-dimensional state encoding classifier confidence, anomaly magnitude, and traffic composition; and an EMA-weighted FedAvg aggregation mechanism that stabilises global model updates in the presence of non-IID client data distributions.

Evaluation on CICDDoS~2019 (215,674 windows, 16~DDoS classes) and TON-IoT (70,335 windows, 10~IoT attack classes) produced the following verified results. On CICDDoS~2019, the framework achieved 99.66\% test accuracy and macro-F1~0.9913 by round~9, converging 81\% faster than DTFL-CD (9 vs.~48 rounds on the same dataset) while matching its minimum 44\% CPU utilisation. A test accuracy of 99.95\% was achieved by the framework on TON-IoT, together with a macro-F1 score of 0.9995 by a round number 10. A best-saved checkpoint at a round number 8 attained a validation macro-F1 score of 0.9998, which outperforms a Fed-Inforce-Fusion baseline by an absolute accuracy margin of 0.55 percentage points while covering three additional attack classes. An individual achievement of an F1 score of at least 0.9989 was observed for all ten TON-IoT classes, including a perfect F1 score of 1.000 for a heavily underrepresented \textit{mitm} class on a balanced test partition. A total of 1,053 and 1,009 verified captures were accumulated by a honeypot system over ten rounds, with an average precision of 0.514 and 0.486 respectively. A confirmation from an explainability analysis is that a SHAP attribution and a LIME attribution align with a domain-known discriminative feature, a Grad-CAM saliency correctly localises an attack-onset timestep, and a counterfactual analysis identifies a minimal feature perturbation that is consistent with a known attack mechanism.

Several acknowledged limitations of the framework exist, and these limitations define an agenda for a future work. A first limitation is that an evaluation used a curated benchmark dataset rather than a live hospital traffic; a real deployment will encounter a device-specific traffic idiosyncrasy, an unrecognised protocol variant, and a long-tail attack type that is not represented in these benchmarks. A second limitation is that a three-client setup represents a minimal federated scenario; a scaling to tens or hundreds of geographically dispersed hospital networks may require a hierarchical aggregation or a client selection strategy. A third limitation concerns a round number 6 oscillation on TON-IoT, which, although dampened by an EMA weighting, was not fully eliminated; an extension of an EMA window ($\alpha \rightarrow 0.2$) or an application of a FedProx proximal penalty to a local TAE update may be necessary for a deployment with a more extreme non-IID partitioning. A fourth limitation is that a current TAE operates on a pre-extracted flow statistic; a future work will explore whether an operation on a raw packet representation or a supplementation of a flow feature with a device-level OS telemetry can further improve a detection of a stealthy attack. A fifth limitation is that all experiments were conducted with a single fixed random seed; a multi-seed evaluation with at least three seeds is recommended to report a mean $\pm$ a standard deviation on a primary metric and to provide a stronger statistical evidence for an observed improvement. A final limitation is that a differential privacy mechanism described in Section~\ref{sec:resilience} was not applied in the reported experiments; an evaluation of an accuracy--privacy trade-off at a different $\epsilon$ value is an important direction for a future work.


\end{document}